\documentclass[final,3p,times,twocolumn]{elsarticle}

\usepackage{amssymb}
\usepackage{amsmath}

\usepackage{booktabs}%
\usepackage{url}
\usepackage{subfig}
\usepackage{algorithmic}
\usepackage[ruled,linesnumbered]{algorithm2e}
\usepackage[table]{xcolor}

\journal{Signal Processing: Image Communication}

\begin{document}

\begin{frontmatter}



\title{Natural Image Stitching using Depth Maps}


\author[1,2]{Tianli Liao}
\ead{tianli.liao@haut.edu.cn}

\author[3,4]{Nan Li\corref{cor1}}
\cortext[cor1]{Corresponding author.}
\ead{nan.li@szu.edu.cn}

\address[1]{Key Laboratory of Grain Information Processing and Control(Henan University of Technology), Ministry of Education, Zhengzhou, 450001, China}
\address[2]{College of Information Science and Engineering, Henan University of Technology, Zhengzhou, 450001, China}
\address[3]{School of Mathematical Sciences, Shenzhen University, Shenzhen, 518060, China}
\address[4]{Guangdong Key Laboratory of Intelligent Information Processing, Shenzhen, 518060, China}


\begin{abstract}
Natural image stitching aims to create a single, natural-looking mosaic from overlapped images that capture the same 3D scene from different viewing positions. Challenges inevitably arise when the scene is non-planar and captured by handheld cameras since parallax is non-negligible in such cases. In this paper, we propose a novel image stitching method using depth maps, which generates accurate alignment mosaics against parallax. Firstly, we construct a robust fitting method to filter out the outliers in feature matches and estimate the epipolar geometry between input images. Then, we utilize epipolar geometry to establish pixel-to-pixel correspondences between the input images and render the warped images using the proposed optimal warping. In the rendering stage, we introduce several modules to solve the mapping artifacts in the warping results and generate the final mosaic. Experimental results on three challenging datasets demonstrate that the depth maps of input images enable our method to provide much more accurate alignment in the overlapping region and view-consistent results in the non-overlapping region. We believe our method will continue to work under the rapid progress of monocular depth estimation. The source code will be made available soon.
\end{abstract}

%


\end{frontmatter}


\section{Introduction}\label{sec1}

\begin{figure*}
	\centering
	\subfloat[Input images and depth maps]{
		\includegraphics[width=0.24\textwidth]{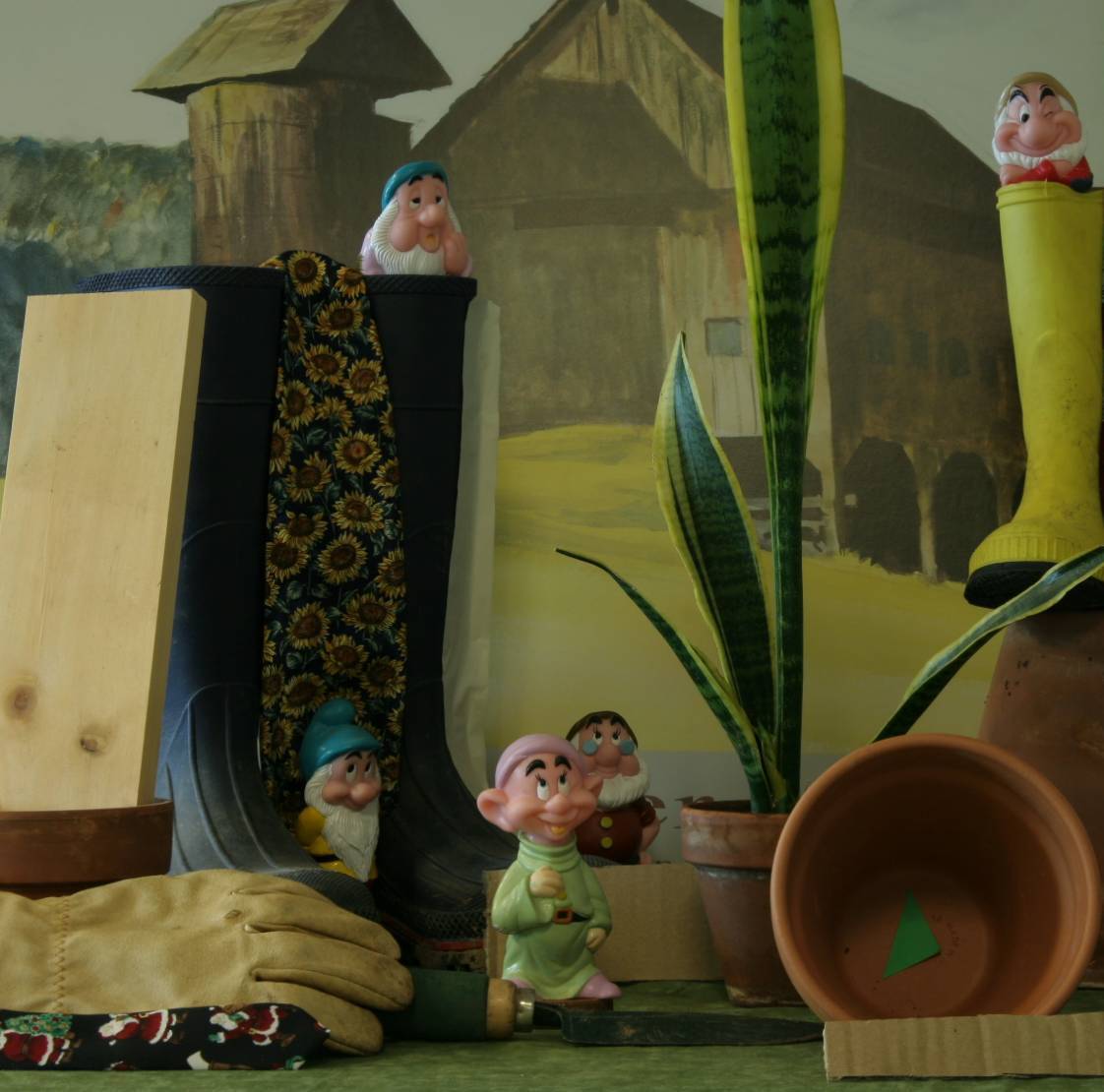}
		\includegraphics[width=0.24\textwidth]{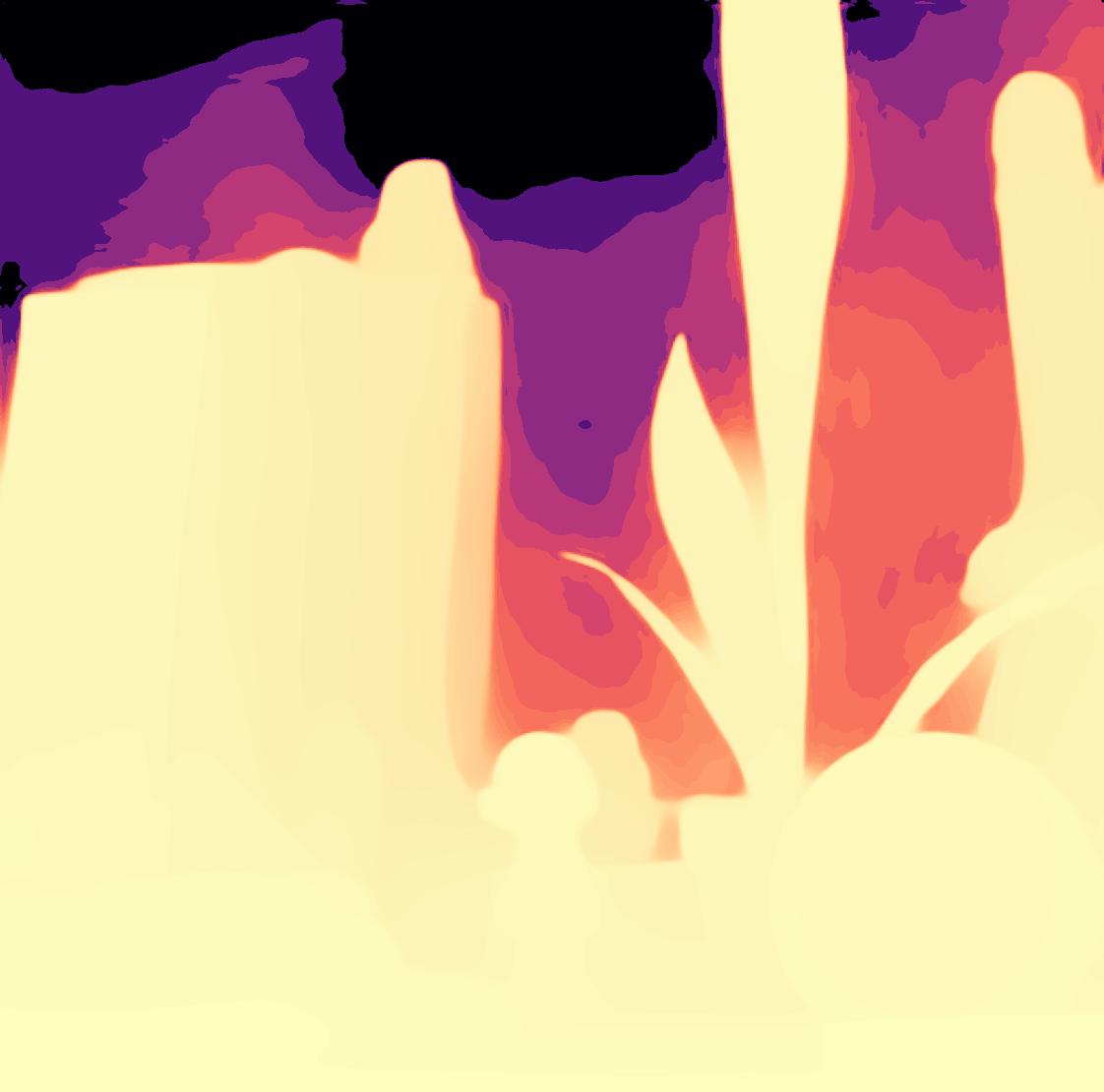}
		\includegraphics[width=0.24\textwidth]{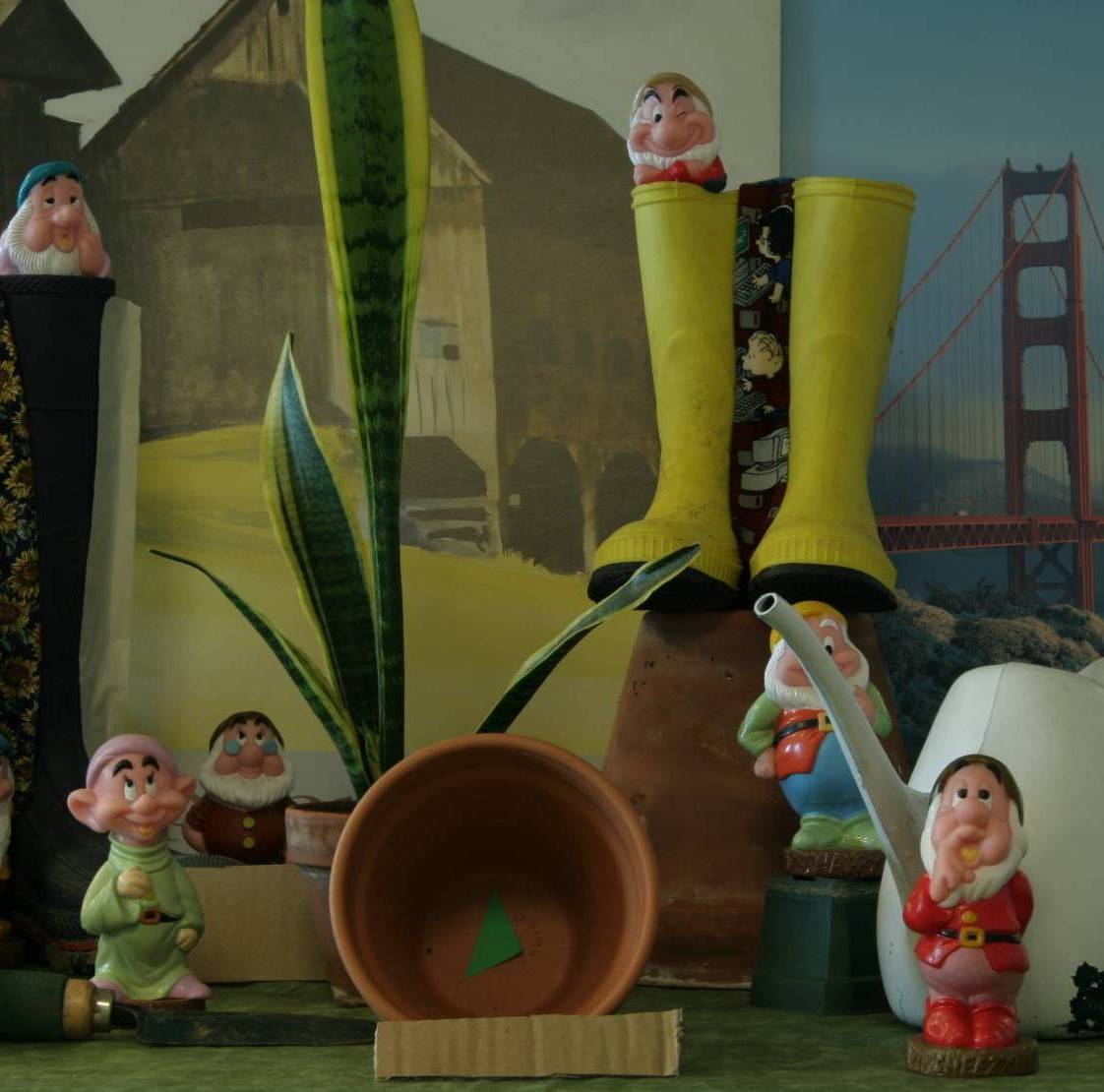}
		\includegraphics[width=0.24\textwidth]{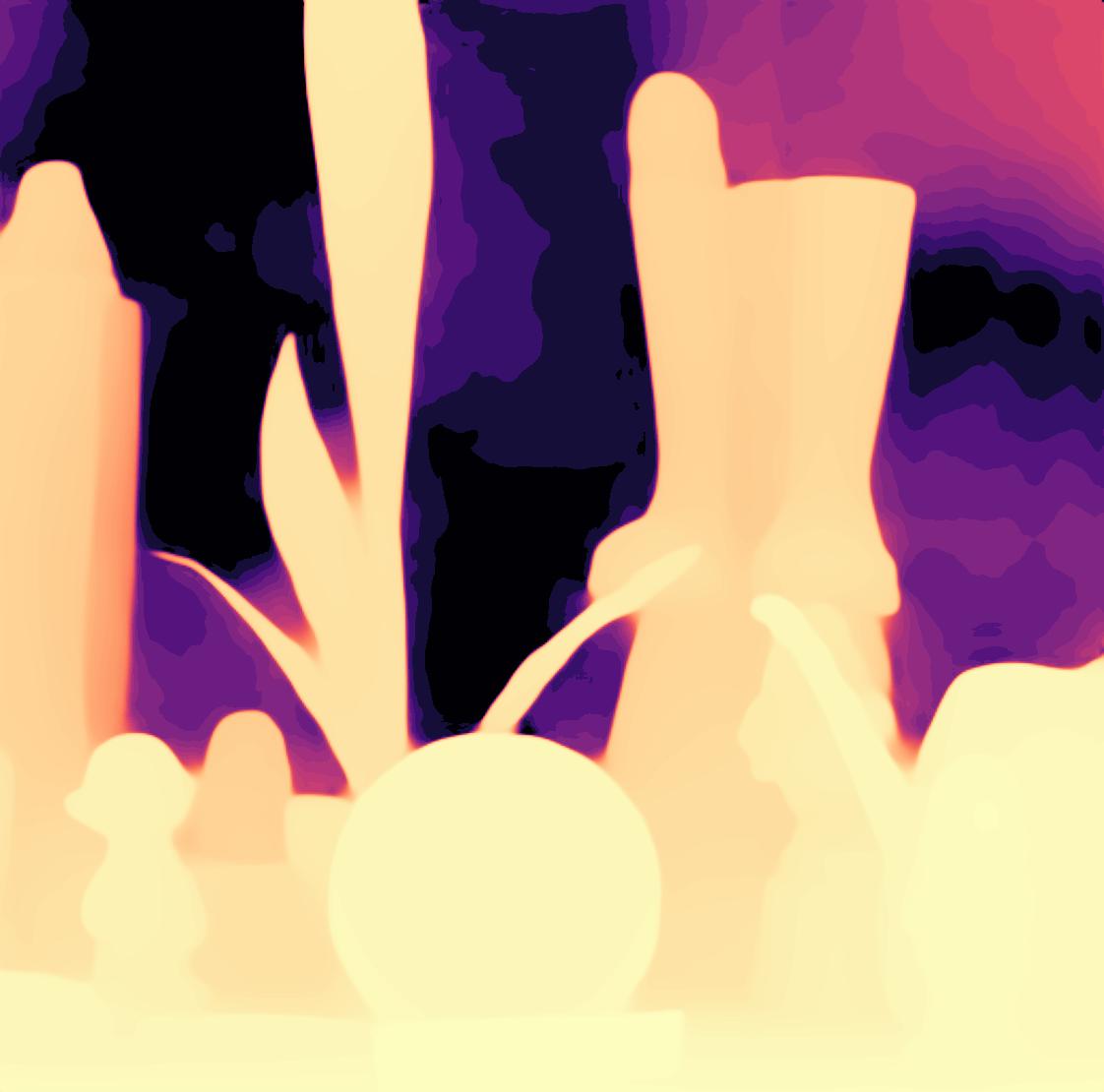}}\\
	\subfloat[Homography]{
		\includegraphics[height=0.125\textheight]{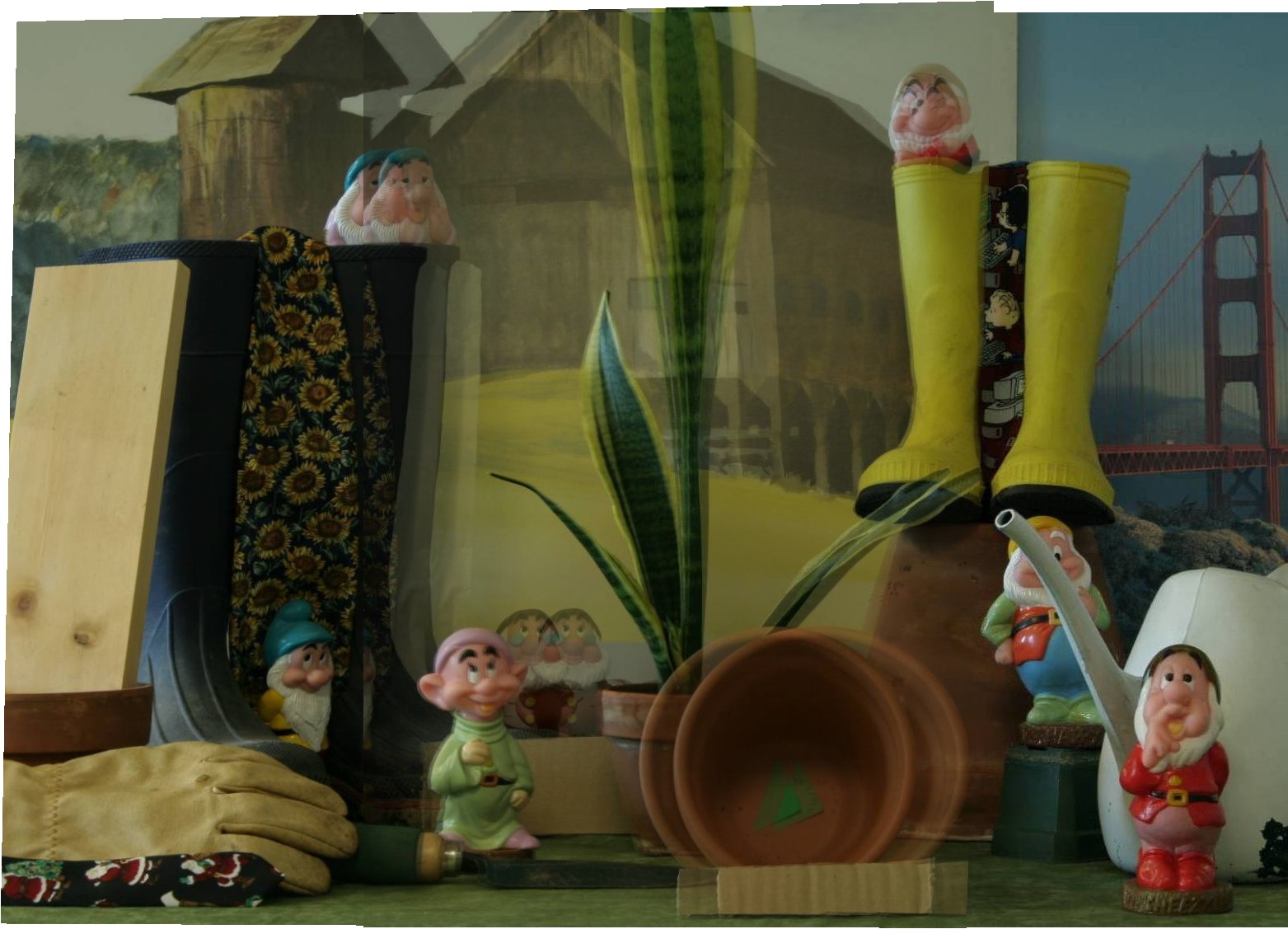}}
	\subfloat[APAP \cite{zaragoza2014projective}]{
		\includegraphics[height=0.125\textheight]{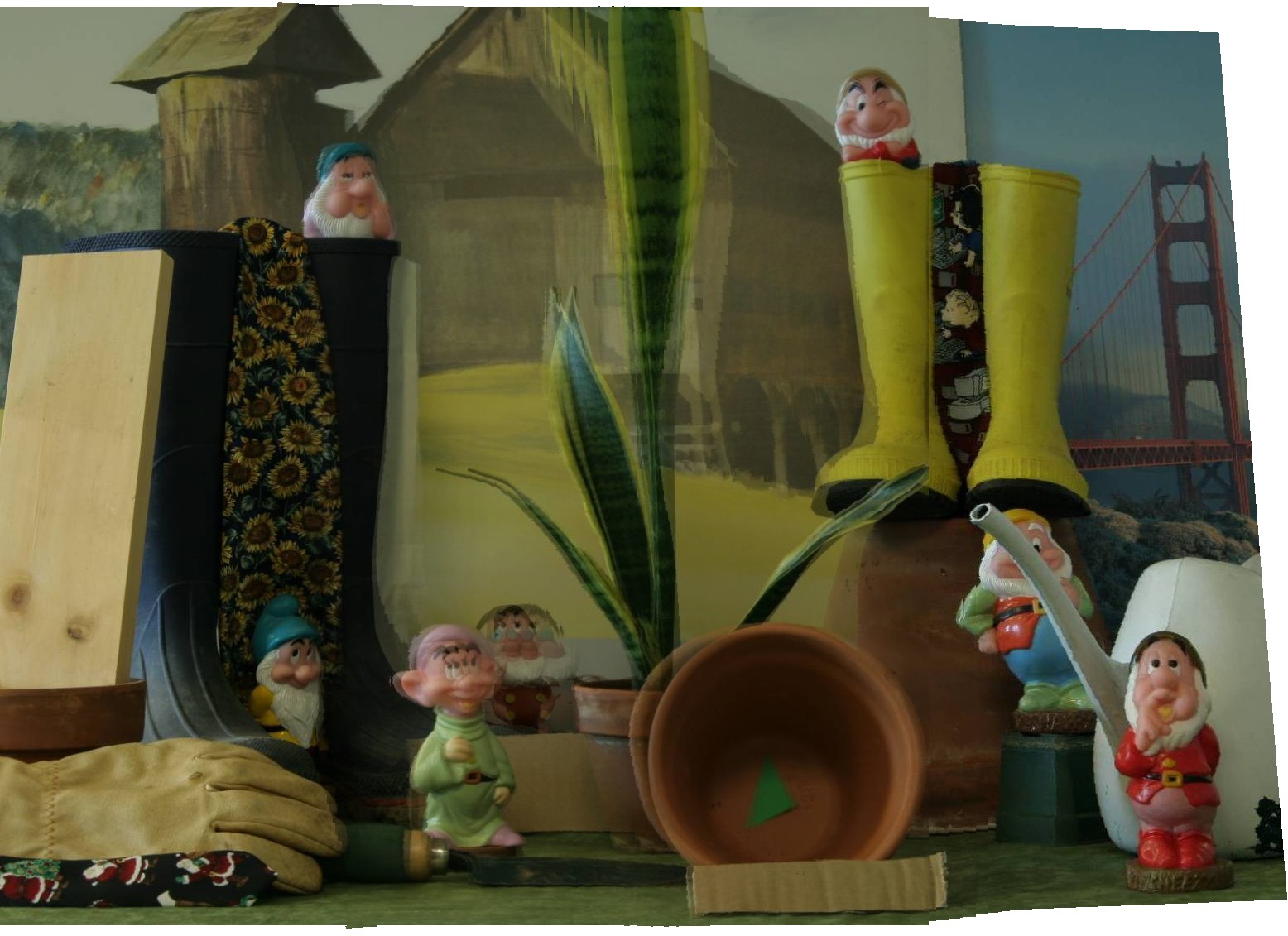}}
	\subfloat[SPHP \cite{chang2014shape}]{
		\includegraphics[height=0.125\textheight]{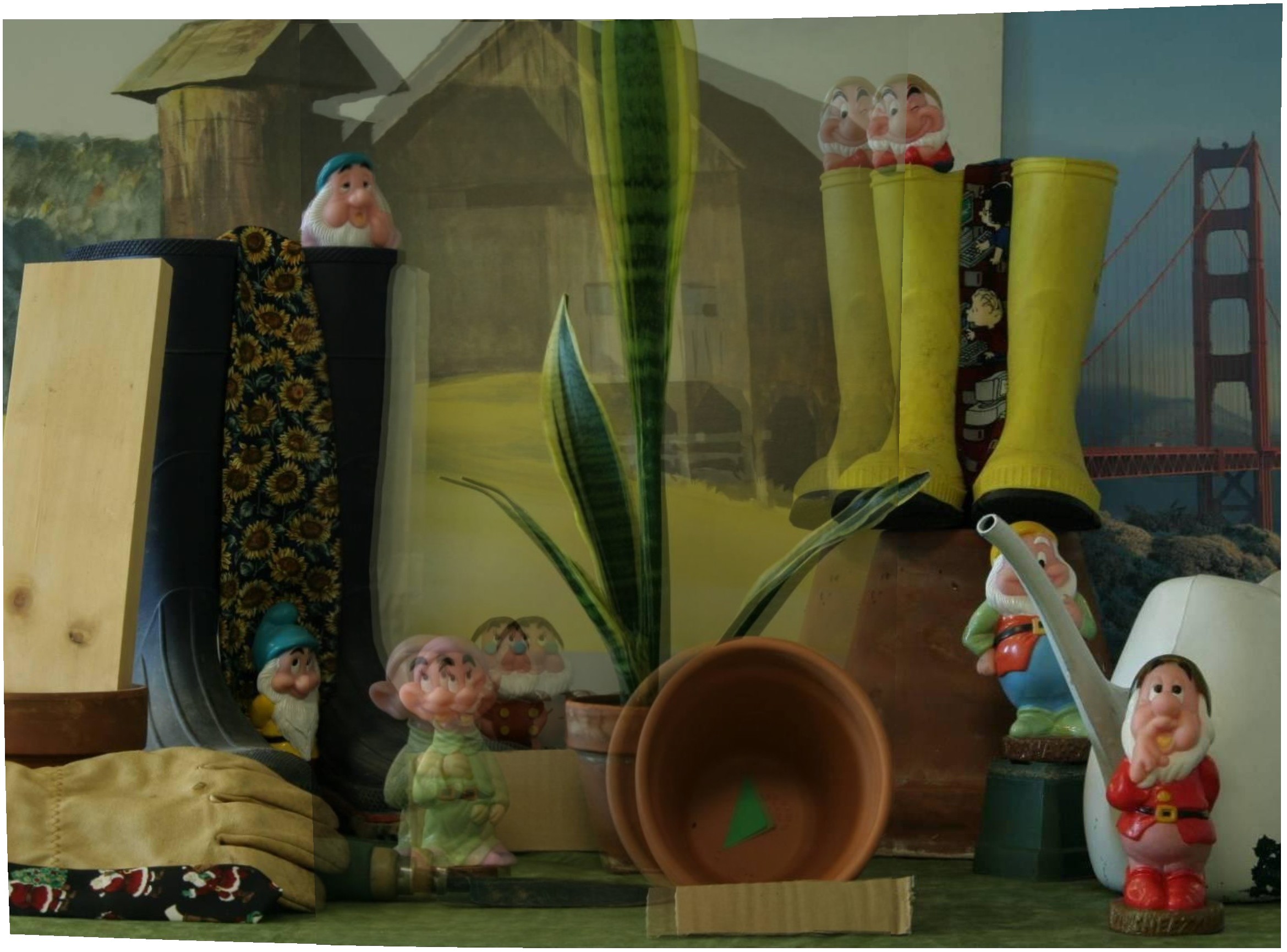}}
	\subfloat[ANAP \cite{lin2015adaptive}]{
		\includegraphics[height=0.125\textheight]{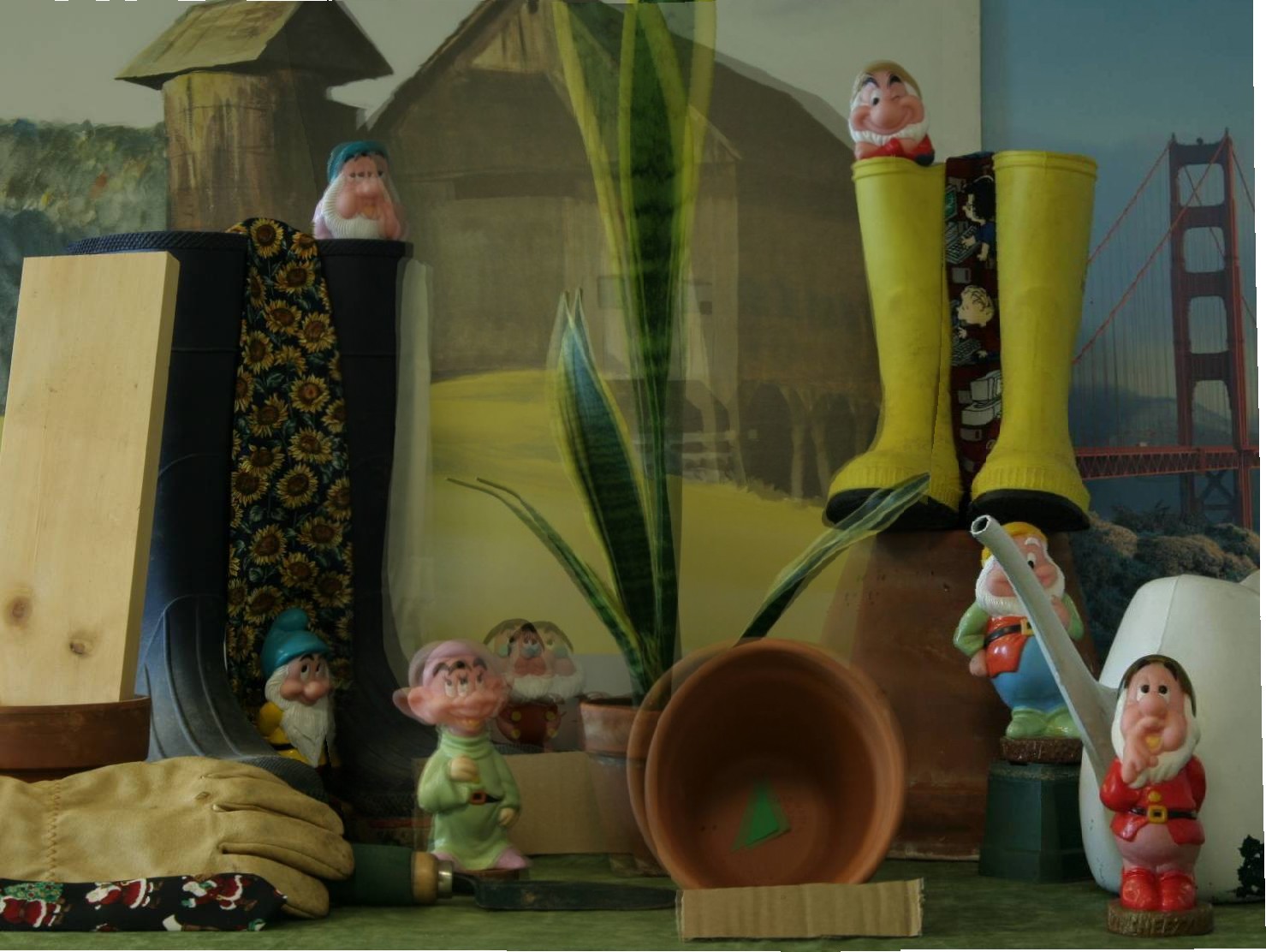}}\\
	\subfloat[GSP \cite{chen2016natural}]{
		\includegraphics[height=0.127\textheight]{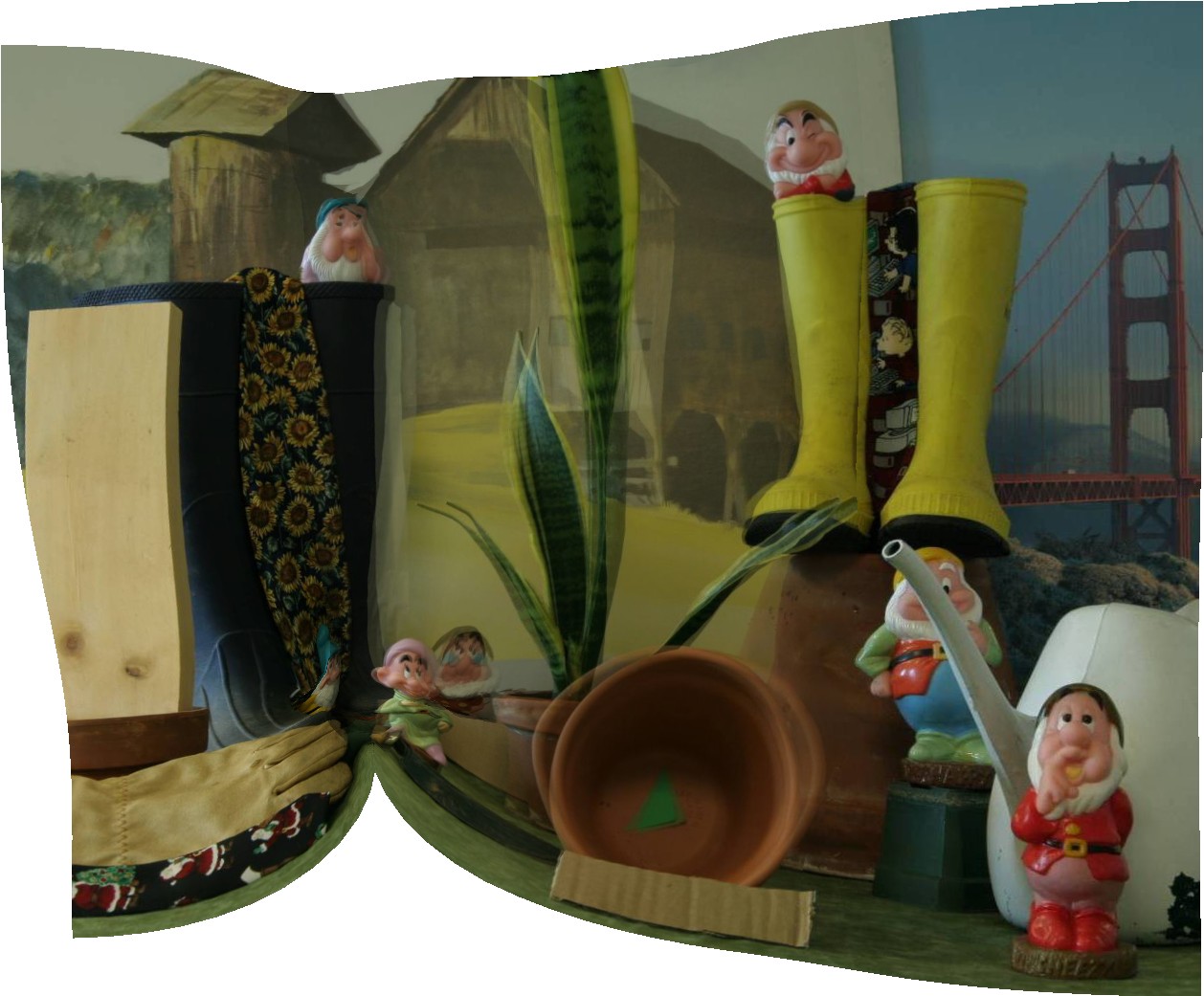}}
	\subfloat[REW \cite{li2018parallax}]{
		\includegraphics[height=0.127\textheight]{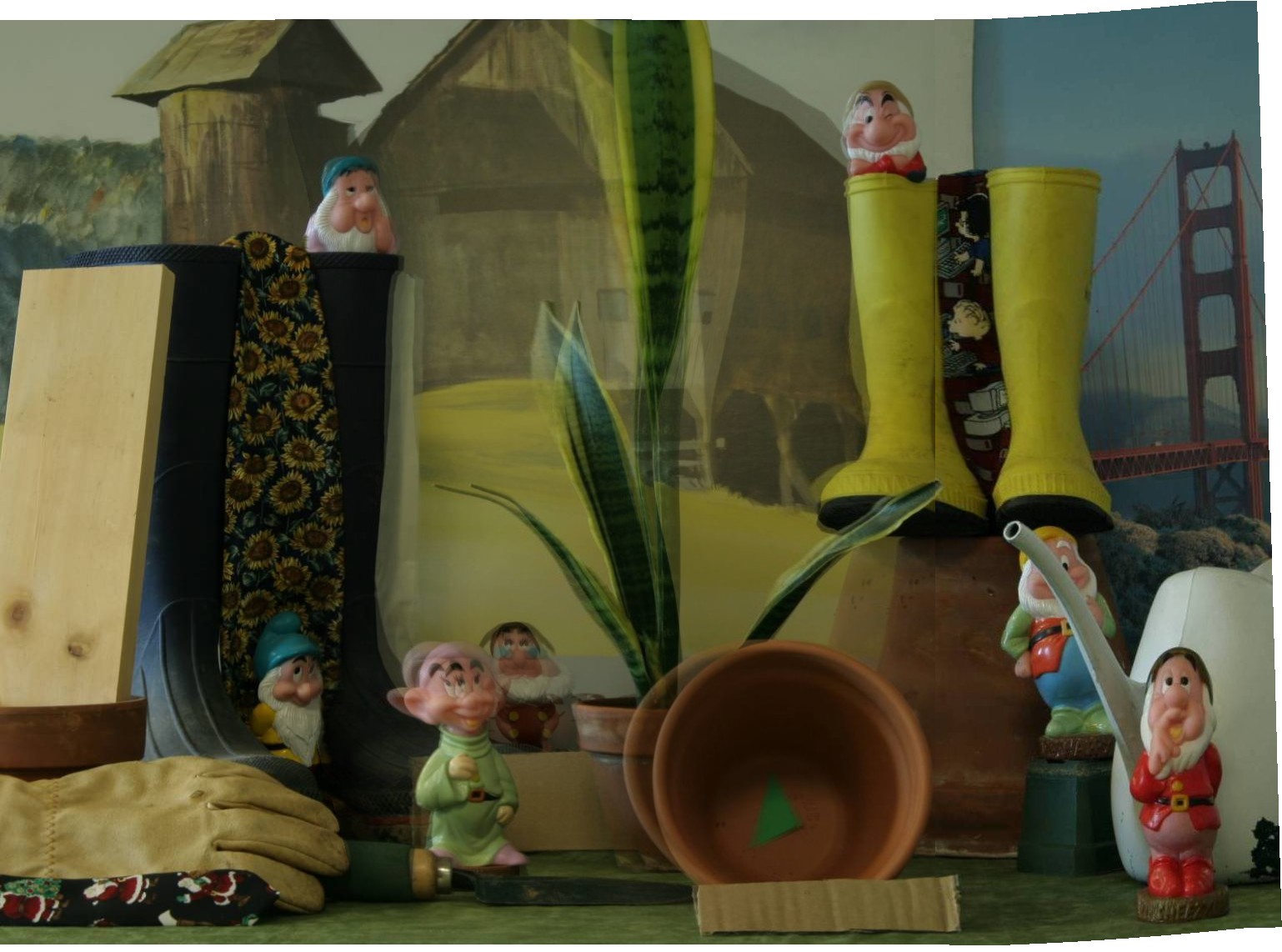}}
	\subfloat[TFA \cite{li2019local}]{
		\includegraphics[height=0.127\textheight]{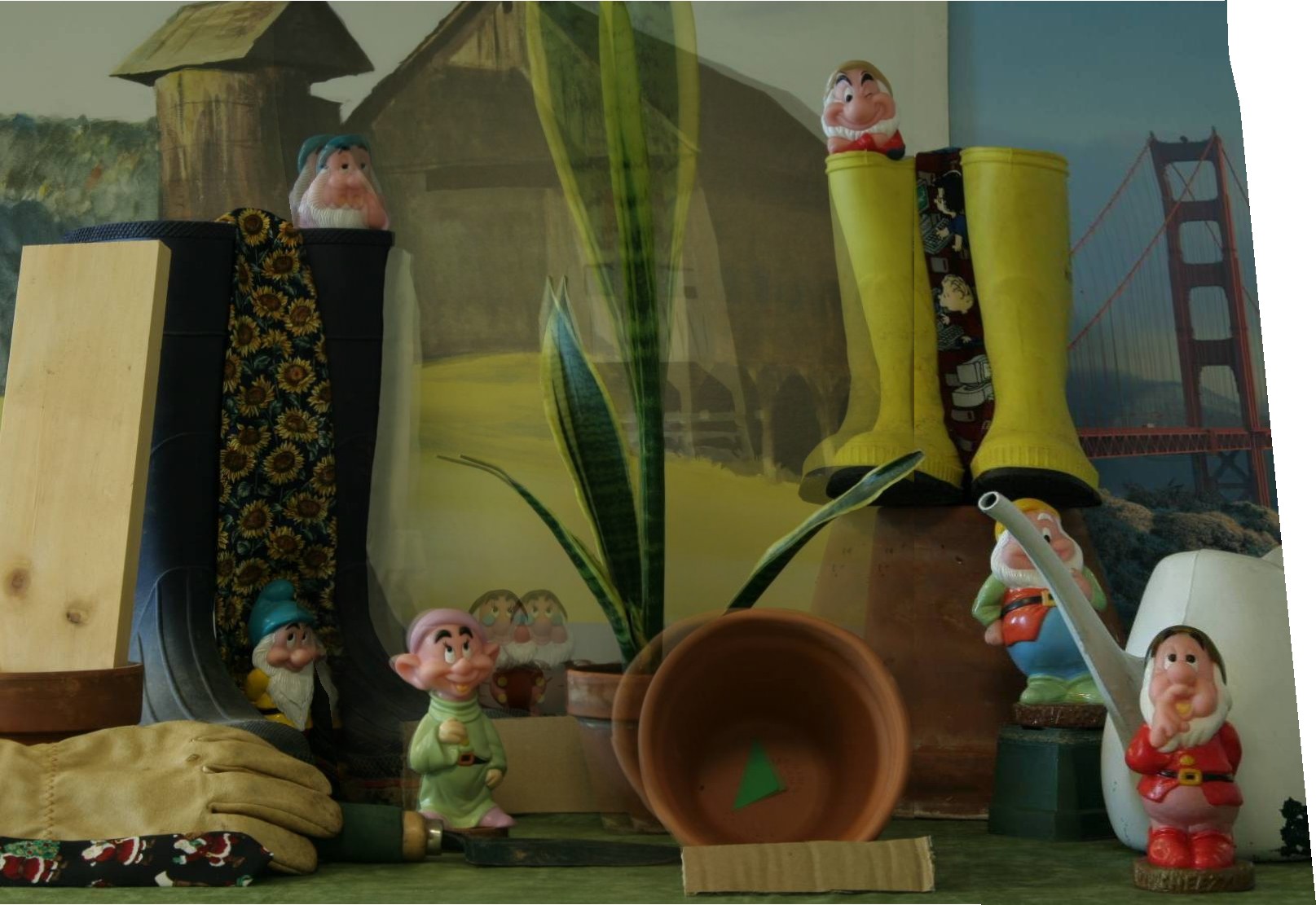}}
	\subfloat[LPC \cite{jia2021Leveraging}]{
		\includegraphics[height=0.127\textheight]{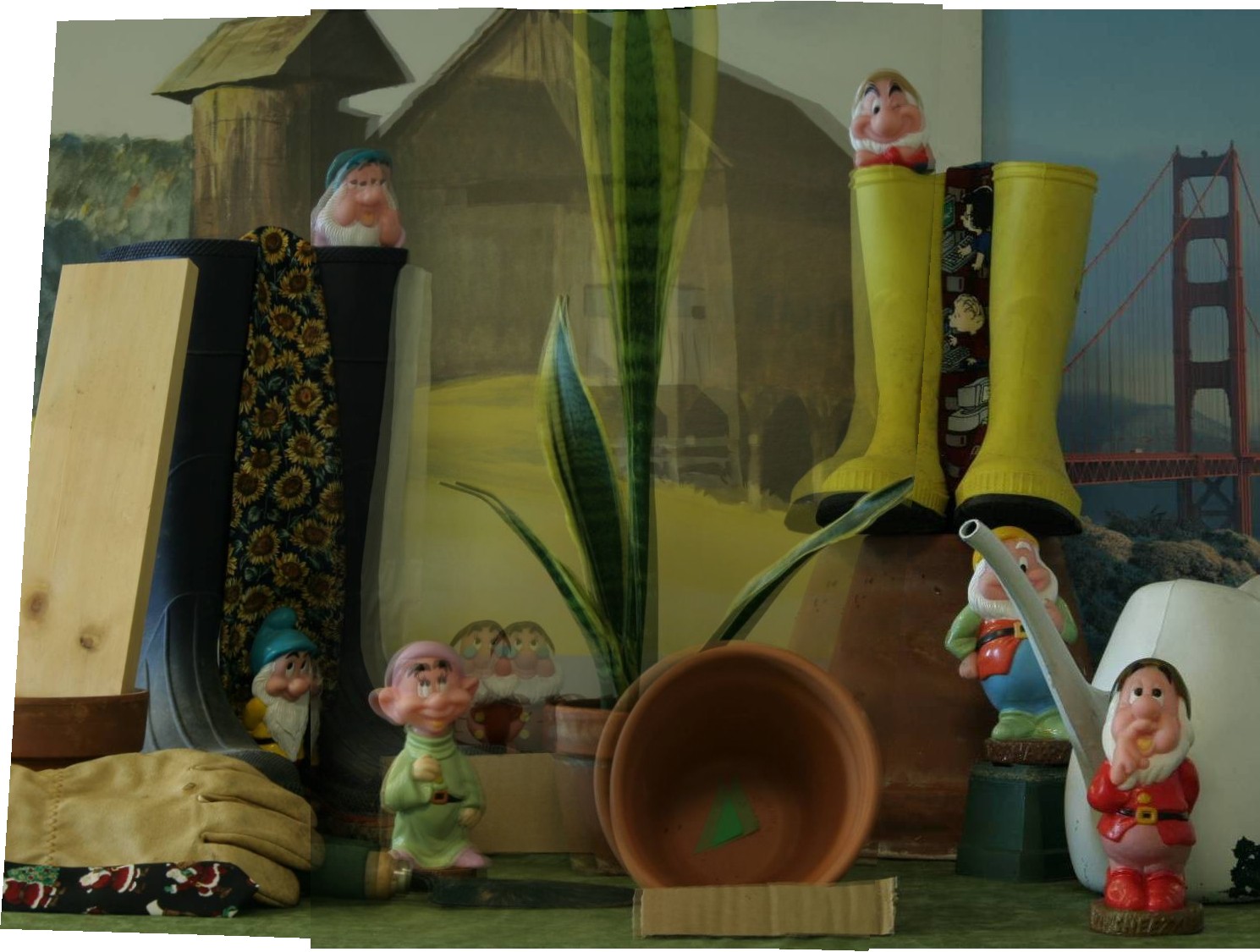}}\\
	\vspace{-0.3cm}
	\subfloat[UDIS++ \cite{nie2023parallax}]{	
		\includegraphics[height=0.12\textheight]{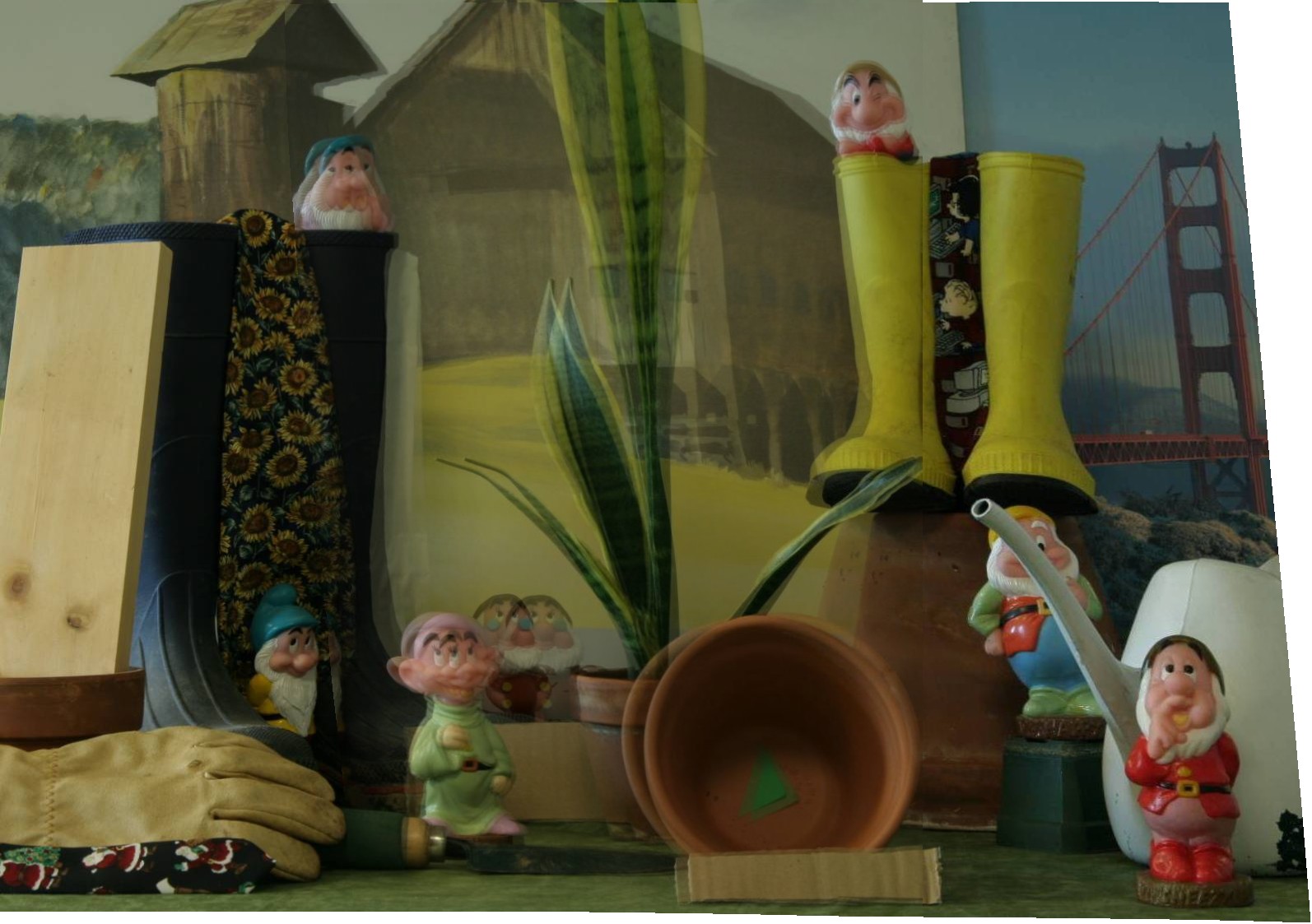}}
	\subfloat[MHW \cite{liao2025parallax}]{		
		\includegraphics[height=0.12\textheight]{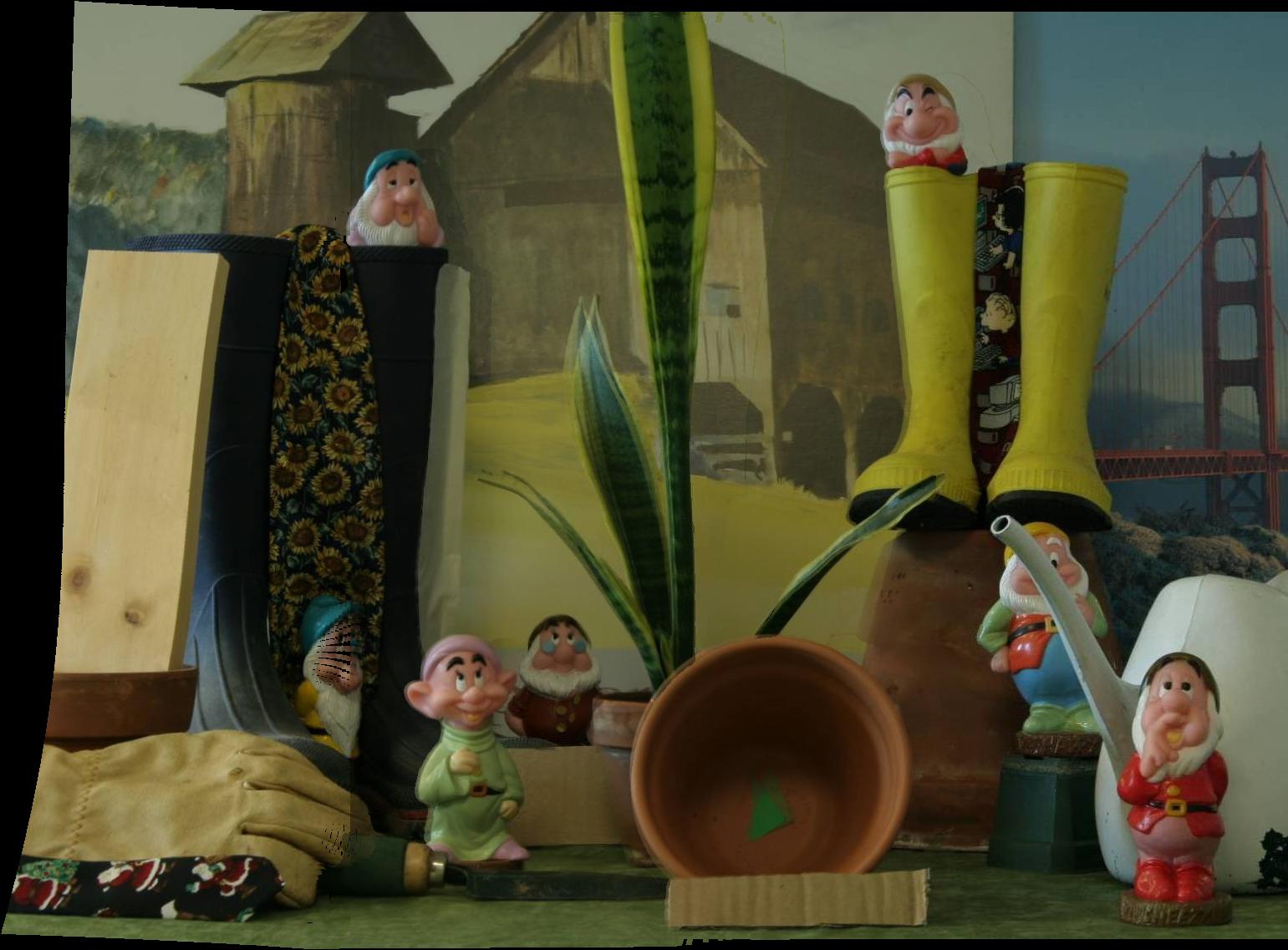}}	
	\subfloat[Warped target image]{
		\includegraphics[height=0.12\textheight]{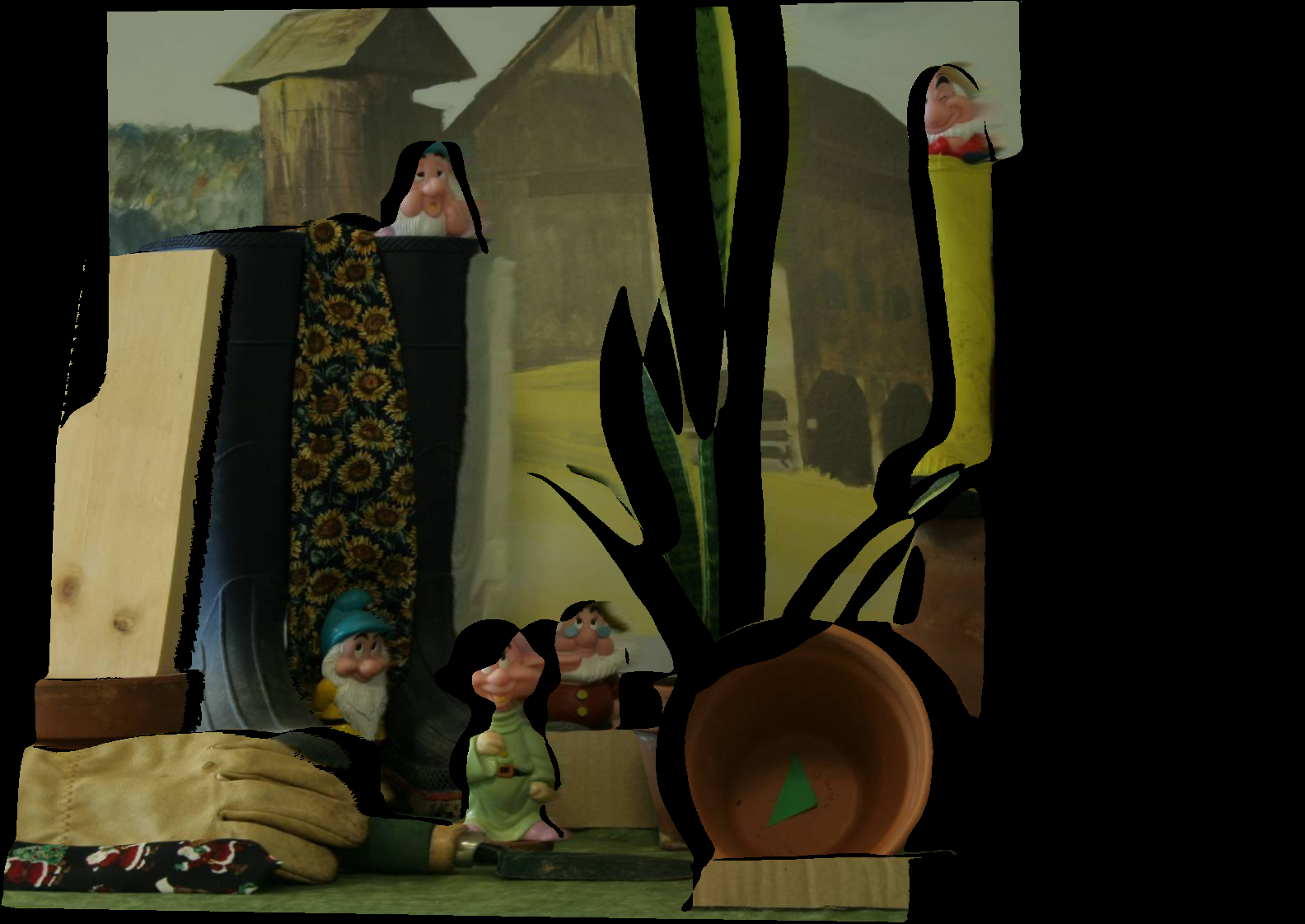}}
	\subfloat[Our final result]{
		\includegraphics[height=0.12\textheight]{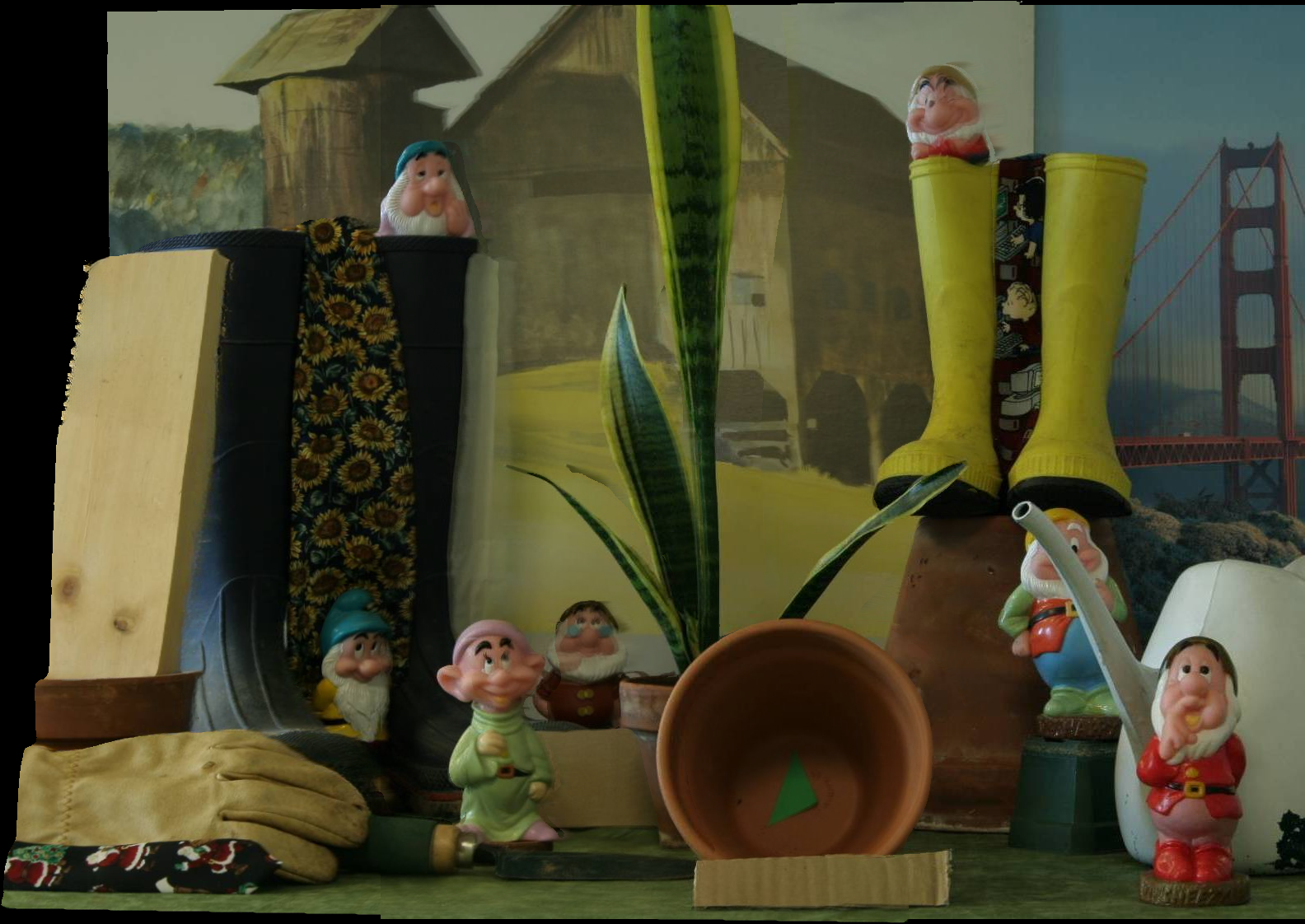}}\\
	\caption{Stitching results of one test case in dataset \cite{zhang2014parallax} via various methods. All results are generated via simple average blending, except that (l) is the warped target image via our method (best view in color and zoom in).}
	\label{fig:1}
\end{figure*}

Natural image stitching is a well-studied problem in computer vision with widespread applications such as video surveillance, autonomous driving, and virtual reality. It aims to composite multiple overlapping images captured from different viewing positions into a single natural-looking panorama \cite{szeliski2006image}. The fundamental problem is $2$-into-$1$: given two input images, one reference and one target, to generate one output image that is virtually captured in the reference viewing position, which includes both overlapping and non-overlapping contents as natural as possible. Hence, the first crucial task is how to warp the target image into an extended view of the reference image, such that the warping result is not only \emph{content-consistent} in the overlapping region but also \emph{view-consistent} in the non-overlapping region. 

When the capturing scene is planar or the viewing point is stationary, homography is effective for accomplishing the dual task \cite{Hartley2004}. However, when the 3D scene consists of background objects with non-planar surfaces or even foreground objects with discontinuous depths, homography cannot generate a plausible mosaic because it is not flexible enough to describe the underlying 3D geometry between parallax views, as shown in Fig. \ref{fig:1}(b).

Lots of adaptive warping models are widely used to address the parallax issue in image stitching. Some methods divide the target image into adjacent patches (pixels \cite{gao2011constructing}, superpixels \cite{Lee_2020_CVPR}, rectangles \cite{zaragoza2014projective}, triangles \cite{li2019local}, irregular domains \cite{Zheng2019tmm}, segments \cite{liao2025parallax}) and warp each of them by a local homography using weighted feature matches; 
some methods divide the target into rectangular cells and deform them simultaneously via an energy minimization 
using local (similar \cite{zhang2014parallax} or affine \cite{zhang2016multi}) plus global (similar \cite{chen2016natural} or linearized projective \cite{liao2020Single,jia2021Leveraging}) geometric invariants. 
Other methods devote attention to combining weighted matches and geometric invariants \cite{chang2014shape,lin2015adaptive,li2017quasi}, increasing densities of feature matches \cite{li2015dual,lin2017direct}, pursuing local alignment allowing seamless composition \cite{gao2013seam,lin2016seagull,zhang2024multimodel}, and formulating image stitching as a learning-based image warping method \cite{nie2020view,nie2021unsupervised,nie2023parallax,jia2023learning}. 
Nevertheless, existing adaptive warping models are still not fine enough to describe the underlying geometry between large-parallax views, such that they still create misaligned mosaics at times (see Fig. \ref{fig:1}(c-k)). 
The inherent reason is that in the presence of large parallax, points lie on multiple depth planes, and no single homography can model all correspondences. Each plane requires its own homography, and the extrapolation of these homographies differs in non-overlapping regions, leading to warping inconsistencies between overlapping and extrapolated non-overlapping regions. Moreover, the number and configuration of such homographies depend on the scene, which makes the model unstable and scene-dependent.

It is well-known that depth maps are powerful for representing the 3D geometry of a stereo scene and can promote stereo vision algorithms with better performance. 
Depth information allows us to recover a simple unified representation of the epipolar geometry that applies to all points in the scene, independent of the magnitude of parallax or scene structure. This not only improves the accuracy of pixel alignment in the overlapping region but also ensures warping consistency across both overlapping and non-overlapping regions, which is not achievable with multiple independent homographies (see Fig. \ref{fig:1}(l,m)).

In this paper, we propose a novel natural image stitching method using depth maps against large parallax. Suppose a set of feature point matches between input images and the depth maps of the images are given, firstly we construct a robust fitting method to filter out the outliers in feature matches and estimate the epipolar geometry of input images, including the infinity homography and epipole; then we construct the pixel-to-pixel correspondences between input images, which are used to render the warped images. In the rendering stage, we propose an optimal warping algorithm, in which several modules are introduced to solve the mapping artifacts in the warped images and generate the final mosaic. Experimental results show that the stitching mosaics by the proposed method can be accurately aligned in the overlapping regions and view-consistent in the non-overlapping regions (see Fig. \ref{fig:1}(m)).

The contributions of our work are as follows:
\begin{itemize}
	\item We propose a robust fitting method to filter out the outliers in feature matches and estimate the epipolar geometry, which is robust to the issue of large parallax;
	\item We propose an image stitching method confirming that depth maps can help provide both \textit{content-consistent} and \textit{view-consistent} results.
\end{itemize}

The rest of the paper is organized as follows. Section \ref{sec:related} reviews the related works of natural image stitching methods. Section \ref{sec:method} proposes the novel method using depth maps. Section \ref{sec:exp} presents the experimental results. Section \ref{sec:conclusion} concludes the paper.

\section{Related Work}
\label{sec:related}

\subsection{Image stitching using piecewise homographies}

Some methods adopted piece-wise homographies as the warping model, where every local homography is determined via some weighting methods. Gao \emph{et al.} \cite{gao2011constructing} proposed a dual-homography warping model, where two representative homographies (distant plane $+$ ground plane) are first clustered, then the local homography per pixel is estimated by a weighted sum of them. Zheng \emph{et al.} \cite{Zheng2019tmm} modified a multiple-homography warping model, where multiple projective-consistent homographies are first clustered and one non-overlapping homography is averaged, then the local homography per pixel is determined by a weighted sum of them. Lin \emph{et al.} \cite{lin2022image} proposed a method using a disparity map and multiple homographies to distinguish one background plane and multiple foreground objects and align them separately. Zhang \emph{et al.} \cite{zhang2024accurate} proposed to estimate multiple warping models for the principal region and then refine the alignment by minimizing pixel-level photometric loss. In our previous work \cite{liao2025parallax}, we proposed to segment images into various contents and estimate multiple homographies to align each content. By using the depth maps, our method can align the images with a single warping model.

Zaragoza \emph{et al.}  \cite{zaragoza2014projective} proposed an as-projective-as-possible (APAP) warp, where the target image is first divided into regular mesh grids and the local homography per mesh is estimated by moving DLT that assigns more weights to feature matches that are located closer to the target mesh. Joo \emph{et al.} \cite{joo2015line} appended line matches into the framework of APAP. Recently, Lee and Sim \cite{Lee_2020_CVPR} proposed a modified version of APAP, where the target image is divided into superpixels instead of meshes, and the local homography per superpixel is estimated by moving DLT, which assigns more weights to feature points that are located on more similar planar regions to the target superpixel instead of explicitly depending on the spatial locations. 

\begin{figure*}
	\centering
	\includegraphics[width=\textwidth]{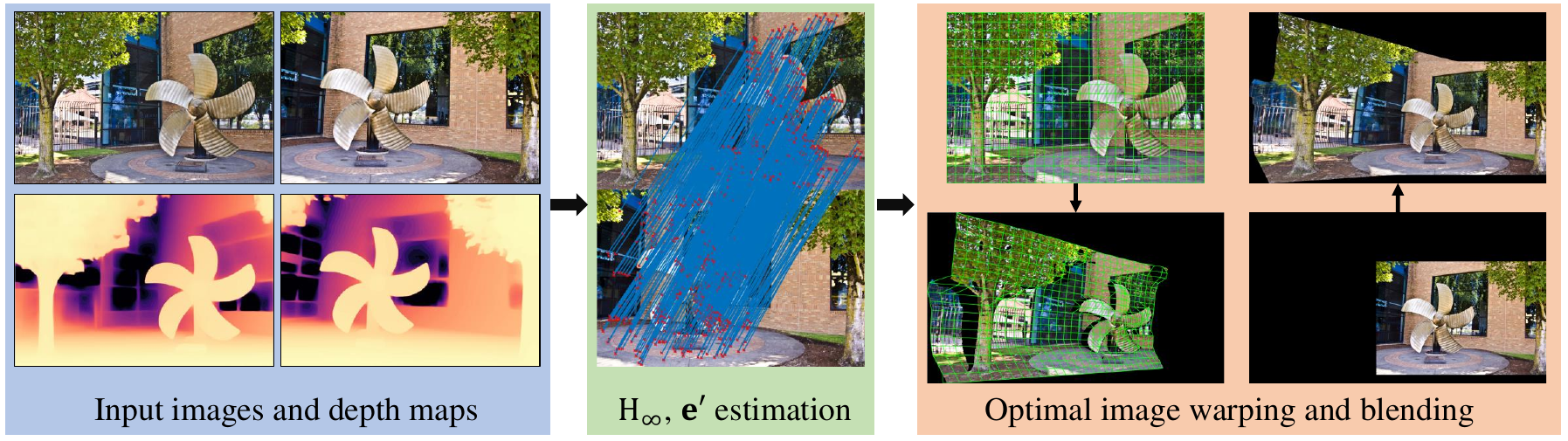}
	\caption{Pipeline of the proposed image stitching method.}
	\label{fig:pipeline}
\end{figure*}

\subsection{Image stitching using geometric invariants}

Instead of using weighted matches to warp non-matching patches, some methods divide the target image into mesh grids and then warp them simultaneously by a deformation, where every mesh is penalized to undergo some geometric invariants (local $+$ global) as much as possible.
Zhang and Liu \cite{zhang2014parallax} proposed a mesh deformation that uses similarity as a local geometric invariant and projective as a global geometric invariant. Chen and Chuang \cite{chen2016natural} used similarity as both local and global geometric invariants. The estimations of global similarity were comprehensively studied in \cite{chang2014shape,lin2015adaptive}.
In order to address the image stitching problem for wide-baseline images, Zhang \emph{et al.} \cite{zhang2016multi} proposed a mesh deformation that uses affine as a local geometric invariant and horizontal-perpendicular-preserving as a global geometric invariant. To generate perspective-consistent mosaics, Liao and Li \cite{liao2020Single} used linearized projective \cite{li2017quasi} as both local and global geometric invariants. Recently, Jia \emph{et al.}  \cite{jia2021Leveraging} proposed a new local coplanar invariant and a new global collinear invariant. Chen \emph{et. al} \cite{chen2021Image} used angle features of key points as geometric invariants to maintain the original rectangular shape of the meshes. Du \emph{et. al} \cite{du2022Geometric} extracted large-scale structures reflected by straight lines or curves as one geometric invariant. Note that local and global geometric invariants play the roles of interpolation and extrapolation regularizers in the overlapping and non-overlapping regions, respectively, while the depth map of the target image can provide a unified and more accurate regularizer.

\subsection{Learning-based methods}

Recent learning-based image alignment and stitching methods can be broadly categorized into homography-based, and continuous warp-based approaches.

Learning-based homography estimation methods~\cite{detone2016deep,le2020deep,nie2020view,zhao2021image,nie2021depth,nie2021unsupervised,nie2022learning,feng2023edge,mei2024dunhuangstitch,li2024dmhomo,wang2024mask} aim to predict single or multiple homographies to align image pairs. These methods perform well for small baseline or near planar scenes but often fail in large parallax or wide baseline scenarios due to their inherent planar assumption.

To overcome this limitation, several studies have proposed continuous warping models such as thin-plate-spline (TPS) motion~\cite{nie2023parallax}, residual elastic warp~\cite{kim2024learning}, or pixel-wise warp~\cite{kweon2023pixel,jia2023learning}. These models enable spatially adaptive transformations and can better accommodate non-planar geometry. However, they remain sensitive to large viewpoint variations and occlusions. 
In contrast, the proposed method targets images with wide baseline and large parallax issue. It leverages depth map  to align images in a single epipolar geometry model, providing an efficient and effective alternative.

\section{Method}
\label{sec:method}

In this section, we propose our method using depth maps, including robust fitting, epipolar geometry estimation, and optimal image warping. The pipeline of our method is shown in Fig. \ref{fig:pipeline}.

\subsection{Robust fitting and epipolar geometry estimation}

\begin{figure*}[t]
	\centering
	\includegraphics[width=0.325\textwidth]{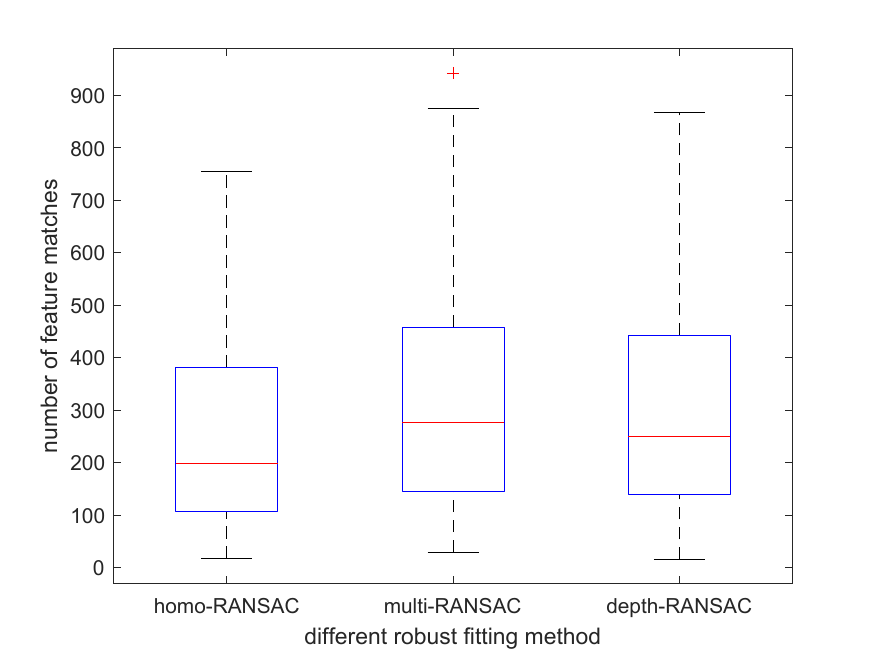}
	\includegraphics[width=0.325\textwidth]{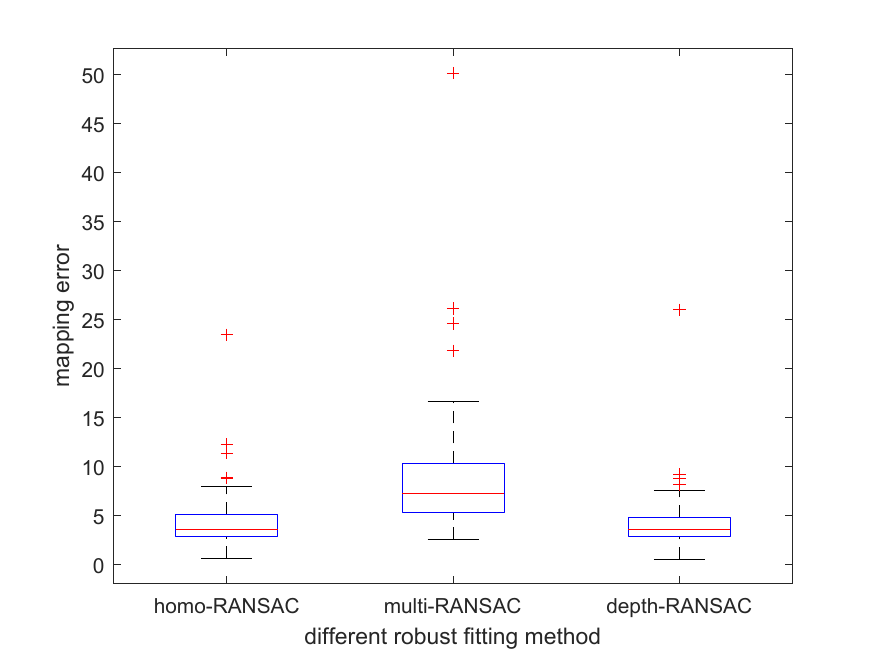}
	\includegraphics[width=0.325\textwidth]{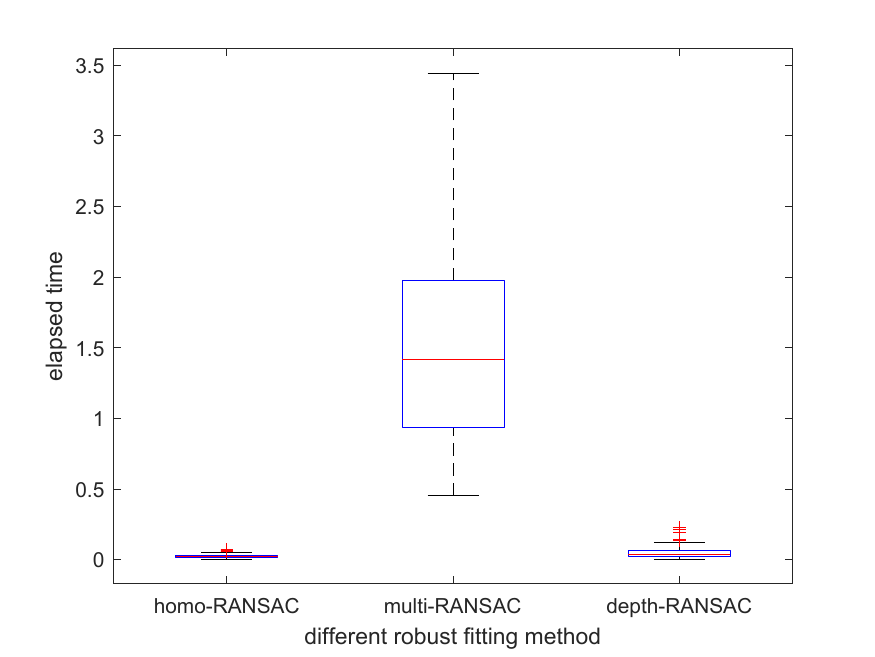}\\
	\caption{Comparison of box plot distributions for different robust fitting methods tested on three datasets \cite{zhang2014parallax,lin2016seagull,herrmann2018robust}. From \textbf{left} to \textbf{right}: The number of feature matches, mapping error, and elapsed time. We test the three methods under different distance threshold settings in RANSAC and record the average values. All the mapping errors are calculated based on Eq. (\ref{eq:maperror}).}
	\label{fig:feature}
\end{figure*}

Given a target image $I_\mathrm{t}$ and a reference image $I_\mathrm{r}$, suppose their camera matrices are:
\begin{equation}\label{camera}
	\mathrm{P}=\mathrm{K}[\mathrm{I} \, |\, \mathbf{0}], \quad \mathrm{P}'=\mathrm{K}'[\mathrm{R}\, |\, \mathbf{t}],
\end{equation}
where $\mathrm{K}\in\mathbb{R}^{3\times3}$ and $\mathrm{K}'\in\mathbb{R}^{3\times3}$ are two calibration matrices, $\mathrm{R}\in\mathrm{SO}(3)$ is a rotation and $\mathbf{t}\in\mathbb{R}^{3}$ is a translation.

Let $\mathbf{X}\in\mathbb{R}^3$ be a world point, $\mathbf{x}\in\mathbb{R}^2$ and $\mathbf{x}'\in\mathbb{R}^2$ be its image points in $I_\mathrm{t}$ and $I_\mathrm{r}$, and $z(\mathbf{x})\in\mathbb{R}$ be its depth value measured from $\mathrm{P}$, then
\begin{equation}\label{projection1}
	\tilde{\mathbf{x}}=\mathrm{K}\mathbf{X}/z(\mathbf{x}).
\end{equation} 
Since $\mathrm{K}$ is invertible, by plugging $\mathbf{X}=z(\mathbf{x})\mathrm{K}^{-1}\tilde{\mathbf{x}}$ into
$\tilde{\mathbf{x}}'\sim\mathrm{K}'\mathrm{R}\mathrm{X}+\mathrm{K}'\mathbf{t}$, we obtain
\begin{equation}\label{depthproj}
	\tilde{\mathbf{x}}' \sim \mathrm{K}'\mathrm{R}\mathrm{K}^{-1}\tilde{\mathbf{x}} + \mathrm{K}'\mathbf{t}/z(\mathbf{x}). 
\end{equation} 
where $\sim$ denotes equality up to scale, the symbol $\tilde{\mathbf{x}}$ (or $\tilde{\mathbf{x}}'$) denotes the homogeneous representation of $\mathbf{x}$ (or $\mathbf{x'}$).

Let $\mathrm{H}_{\infty} = \mathrm{K}'\mathrm{R}\mathrm{K}^{-1}$ and $\mathbf{e}' = \mathrm{K}'\mathbf{t}$, we simplify Eq.~(\ref{depthproj}) as
\begin{equation}\label{Hande}
	\tilde{\mathbf{x}}' \sim \mathrm{H}_{\infty}\tilde{\mathbf{x}} + \mathbf{e}'/z(\mathbf{x}).
\end{equation}
In fact, $\mathrm{H}_{\infty}$ is the infinite homography between two parallax views, and $\mathbf{e}'$ is the epipole in the view of $I_\mathrm{r}$.

If a pair of feature matches is incorrect (an outlier), the mapping error would increase extremely, such that we can construct a robust fitting method based on Eq.~(\ref{Hande}) to filter out the outliers in feature matches. The mapping error of a feature match $(\mathbf{p}_i,\mathbf{q}_i)$ is calculated as
\begin{equation}\label{eq:maperror}
	\epsilon_i = \left\|\pi\left(\mathrm{H}_{\infty}\,\tilde{\mathbf{p}}_i+\frac{\mathbf{e}'}{z(\mathbf{p}_i)}\right)-\mathbf{q}_i\right\|,	
\end{equation}
where $\pi(\mathbf{v})=(v_1/v_3,v_2/v_3)^T$ for $\mathbf{v}=(v_1,v_2,v_3)^T\in\mathbb{R}^3$. Conversely, if a set of inliers and their corresponding depth values from $\mathrm{P}$ are given, one can estimate $\mathrm{H}_{\infty}$ and $\mathbf{e}'$ based on Eq.~(\ref{Hande}).

In order to estimate $\mathrm{H}_{\infty}$ and $\mathbf{e}'$, we firstly prepare a set of SIFT \cite{lowe2004distinctive} point matches $\{(\mathbf{p}_i, \mathbf{q}_i)\}_{i=1}^N$ between $I_\mathrm{t}$ and $I_\mathrm{r}$, a depth map $z=z(\mathbf{x})$ of $I_\mathrm{t}$.
Similar to the DLT algorithm for estimating homography from a data set $\{(\mathbf{p}_i, \mathbf{q}_i)\}_{i=1}^N$, $\mathrm{H}_{\infty}$ and $\mathbf{e}'$ can be estimated from the augmented data set $\{(\mathbf{p}_i, \mathbf{q}_i, z(\mathbf{p}_i))\}_{i=1}^N$ via solving the following linear least-square problem
\begin{equation}
	\min\limits_{\mathbf{h},\mathbf{e}'}~\left\|\mathrm{A}\,\mathbf{h}+\mathrm{B}\,\mathbf{e}'\right\|^2, 
	\label{eq:He1}
\end{equation}
where $\mathbf{h}$ is a $9$-vector made up of the entries of $\mathrm{H}_\infty$. The matrices $\mathrm{A}$ and $\mathrm{B}$ are vertically stacked by
\begin{equation}
	\mathrm{A}_i=\begin{bmatrix}
		x_i & y_i & 1 & 0 & 0 & 0 & -x_i x'_i & -x'_i y_i & -x'_i\\
		0 & 0 & 0 & x_i & y_i & 1 & -x_i y'_i & -y_i y'_i & -y'_i
	\end{bmatrix}
\end{equation}
\begin{equation}
	\mathrm{B}_i=\begin{bmatrix}
		1/z_i & 0 & -x'_i/z_i\\
		0 & 1/z_i & -y'_i/z_i
	\end{bmatrix},
\end{equation}
for $i=1,\ldots,N$, $(x_i,y_i)$ and $(x'_i,y'_i)$ are the coordinates of $\mathbf{p}_i$ and $\mathbf{q}_i$, $z_i=z(\mathbf{p}_i)$.

When $N\geq6$, Eq.~(\ref{eq:He1}) can be efficiently solved by Singular Value Decomposition (SVD). For the sake of more robust and accurate estimation, we employ the $6$-point SVD solver as the minimal solver in the RANSAC framework and calculate $\mathrm{H}_{\infty}$ and $\mathbf{e}'$ by solving the following nonlinear least squares problem
\begin{equation}
	\min_{\mathrm{H}_{\infty},\mathbf{e}'}~\sum_{i\in \mathrm{IS}}\|\pi(\mathrm{H}_{\infty}\,\tilde{\mathbf{p}}_i+\mathbf{e}'/z(\mathbf{p}_i))-\mathbf{q}_i\|^2,
	\label{eq:He2}
\end{equation}
where $\mathrm{IS}$ is the index set of identified inliers from the RANSAC estimator. Eq.~(\ref{eq:He2}) can be efficiently solved by the Levenberg-Marquardt algorithm.	The algorithm for estimating $\mathrm{H}_\infty$ and $\mathbf{e}'$ is summarized in Algorithm \ref{algor_1}.

With the help of depth data, a single RANSAC estimator can identify a sufficiently large consensus set of point matches between large parallax views, while existing methods need multiple RANSAC estimators to identify multiple homographies. 
Fig.~\ref{fig:feature} shows the comparison results of box plot distributions of the number of feature matches, mapping error, and elapsed time via three robust fitting methods: homography-based RANSAC \cite{fischler1981random} (homo-RANSAC), multiple sampling RANSAC \cite{zaragoza2014projective} (multi-RANSAC), and our depth-based RANSAC (depth-RANSAC). The three robust fitting methods are applied to identify the inliers and compute the corresponding epipolar geometry via Eq. (\ref{eq:He2}), respectively.
Our depth-RANSAC method can identify a comparable number of feature matches with multi-RANSAC. Meanwhile, it takes very little time and has the lowest and most stable mapping errors. The reason why multi-RANSAC identifies the largest number of feature matches but has the worst mapping error is that the depth values of some features are inaccurate, which results in an unstable estimation of the epipolar geometry with large errors. Our method is more robust against inaccurate depth estimation.
More experiments on the superiority of our depth-based RANSAC are demonstrated in Sec. \ref{subsec:ablation}.

\begin{algorithm}
	\caption{Estimate $\mathrm{H}_{\infty}$ and $\mathbf{e}'$}
	\label{algor_1} 
		\KwIn{ feature matches $\{(\mathbf{p}_i, \mathbf{q}_i)\}_{i=1}^N$ and depth values $\{z_i=d(\mathbf{p}_i)\}_{i=1}^N$}
		\KwOut{infinite homography $\hat{\mathrm{H}}_{\infty}$ and epipole $\hat{\mathbf{e}}'$}
		initialize $\hat{\mathrm{H}}_{\infty}$, $\hat{\mathbf{e}}'$ and identify inliers via RANSAC with a minimal six-point SVD solver Eq. (\ref{eq:He1});\\
		refine $\hat{\mathrm{H}}_{\infty}$ and $\hat{\mathbf{e}}'$ by optimizing Eq. (\ref{eq:He2});\\
		return $\hat{\mathrm{H}}_{\infty}$ and $\hat{\mathbf{e}}'$. 
\end{algorithm}

\subsection{Optimal image warping}

\begin{figure*}[t]
	\centering
	\subfloat[Reference image]{
		\includegraphics[width=0.305\textwidth]{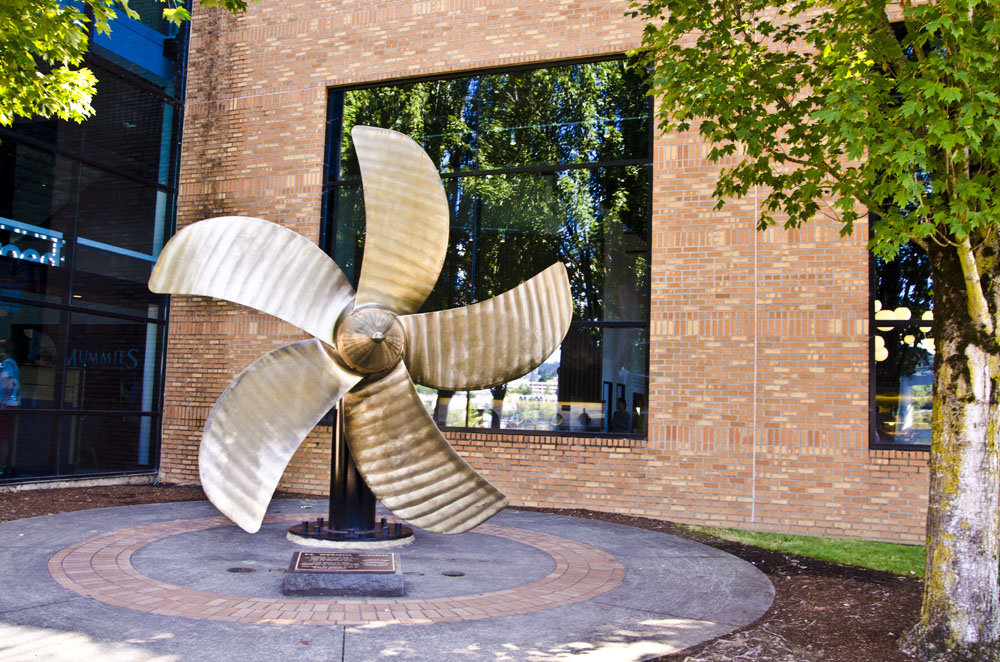}}
	\subfloat[Target image with mesh grids]{
		\includegraphics[width=0.305\textwidth]{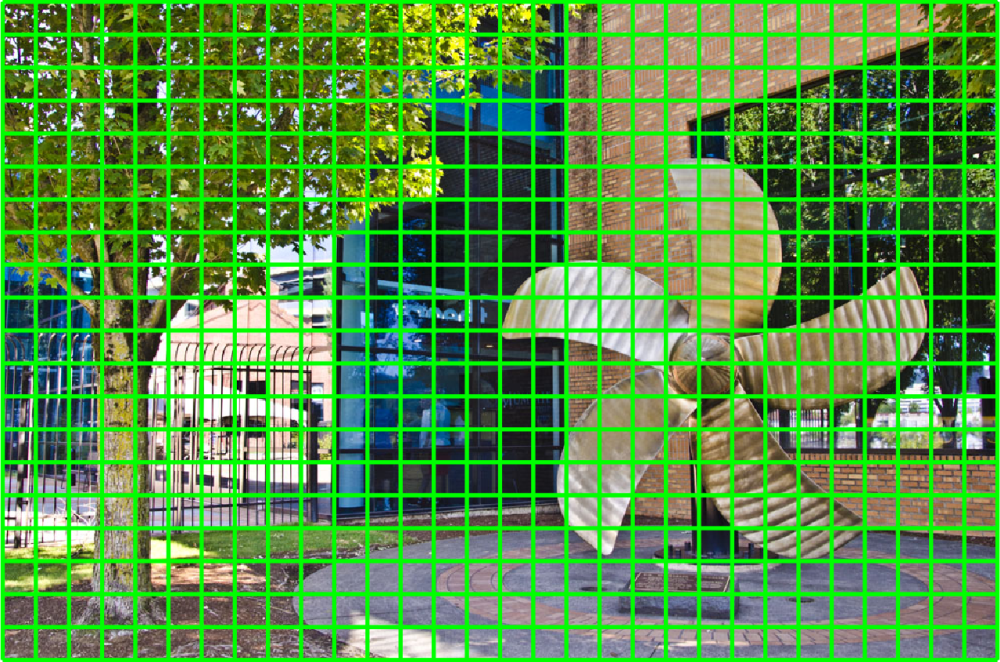}}
	\subfloat[Our result with warped mesh grids]{
		\includegraphics[width=0.35\textwidth]{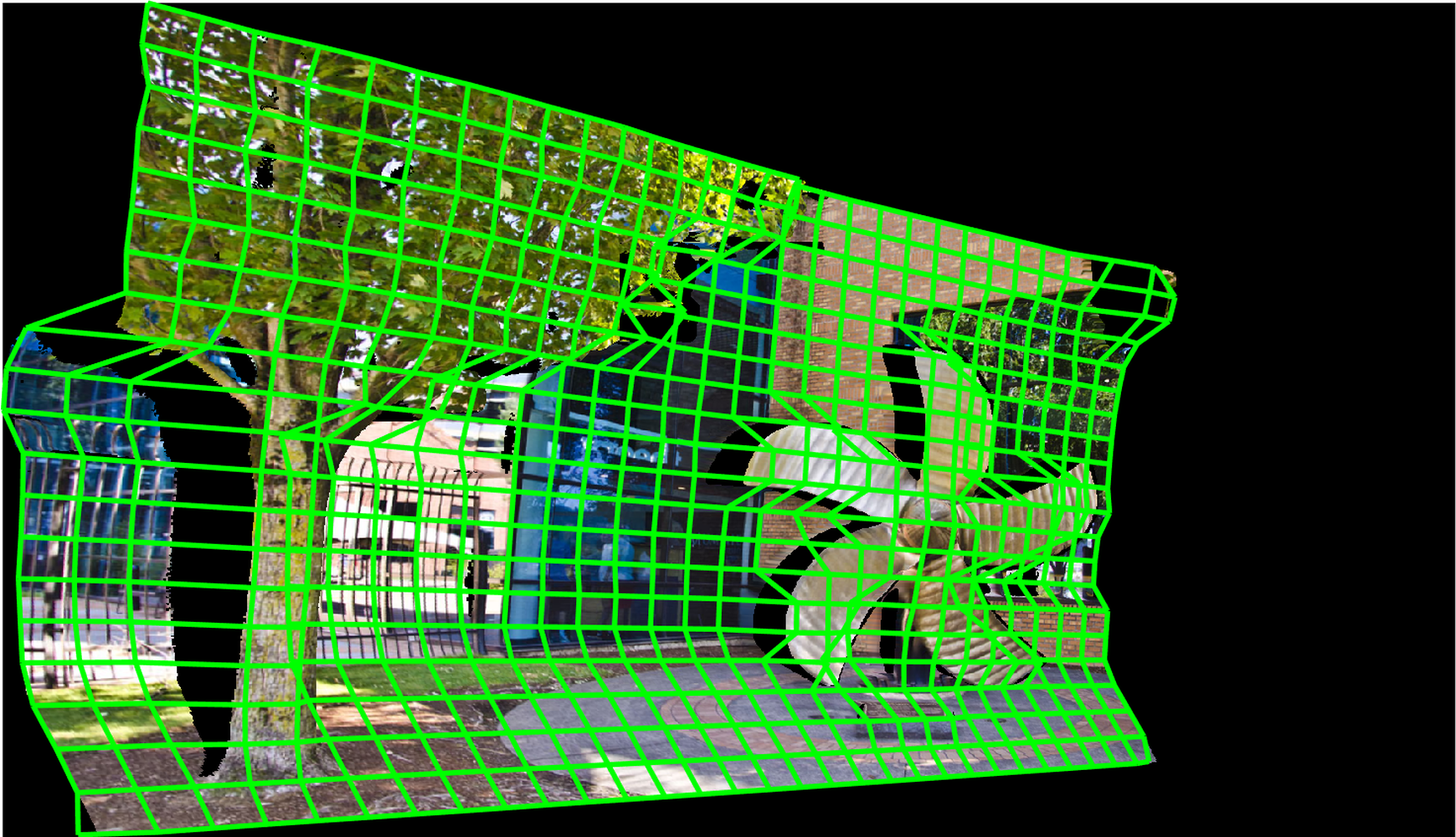}}\\
	\vspace{-0.2cm}
	\subfloat[Forward mapping result]{
		\includegraphics[width=0.32\textwidth]{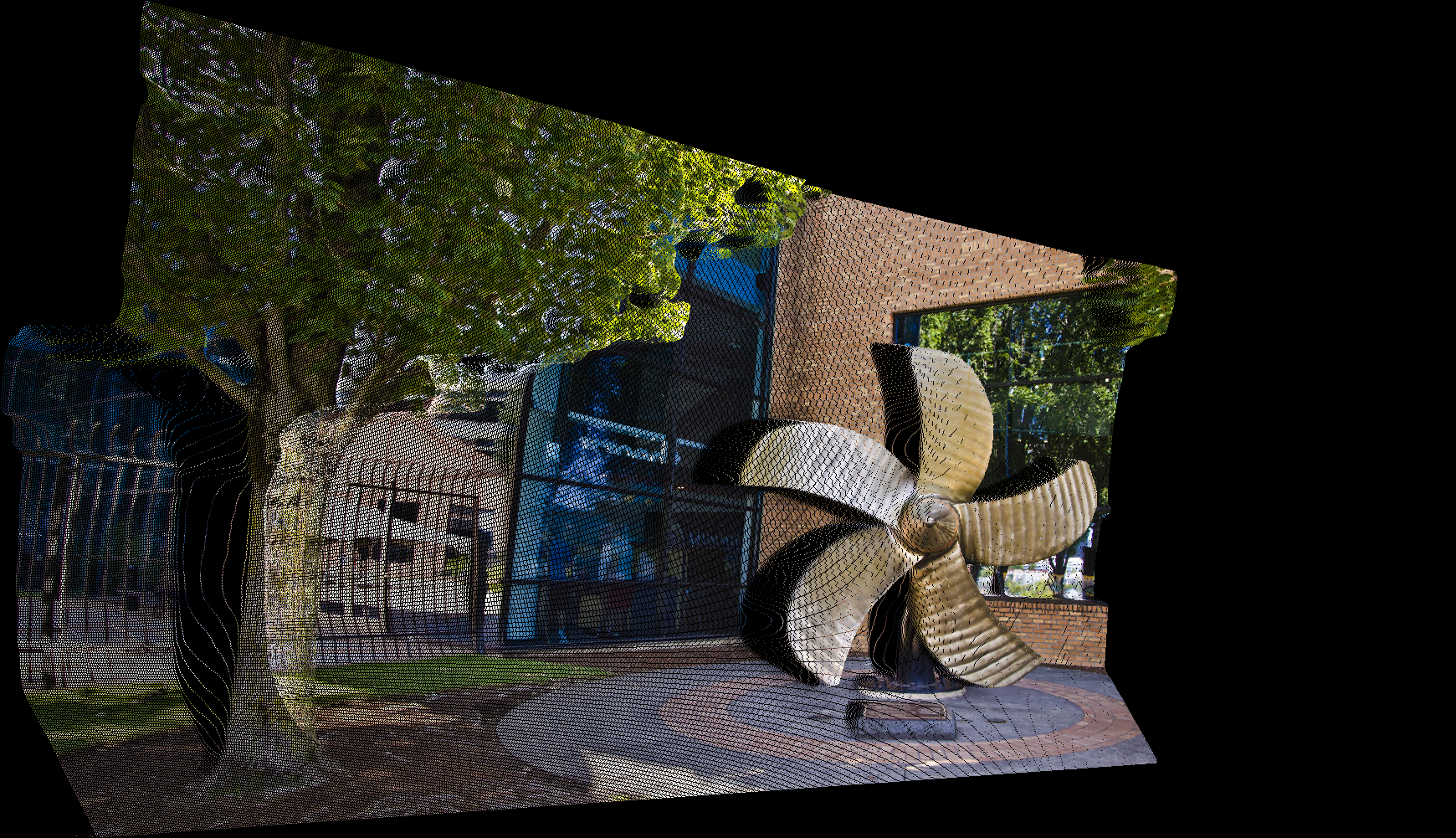}}
	\subfloat[Backward mapping result]{
		\includegraphics[width=0.32\textwidth]{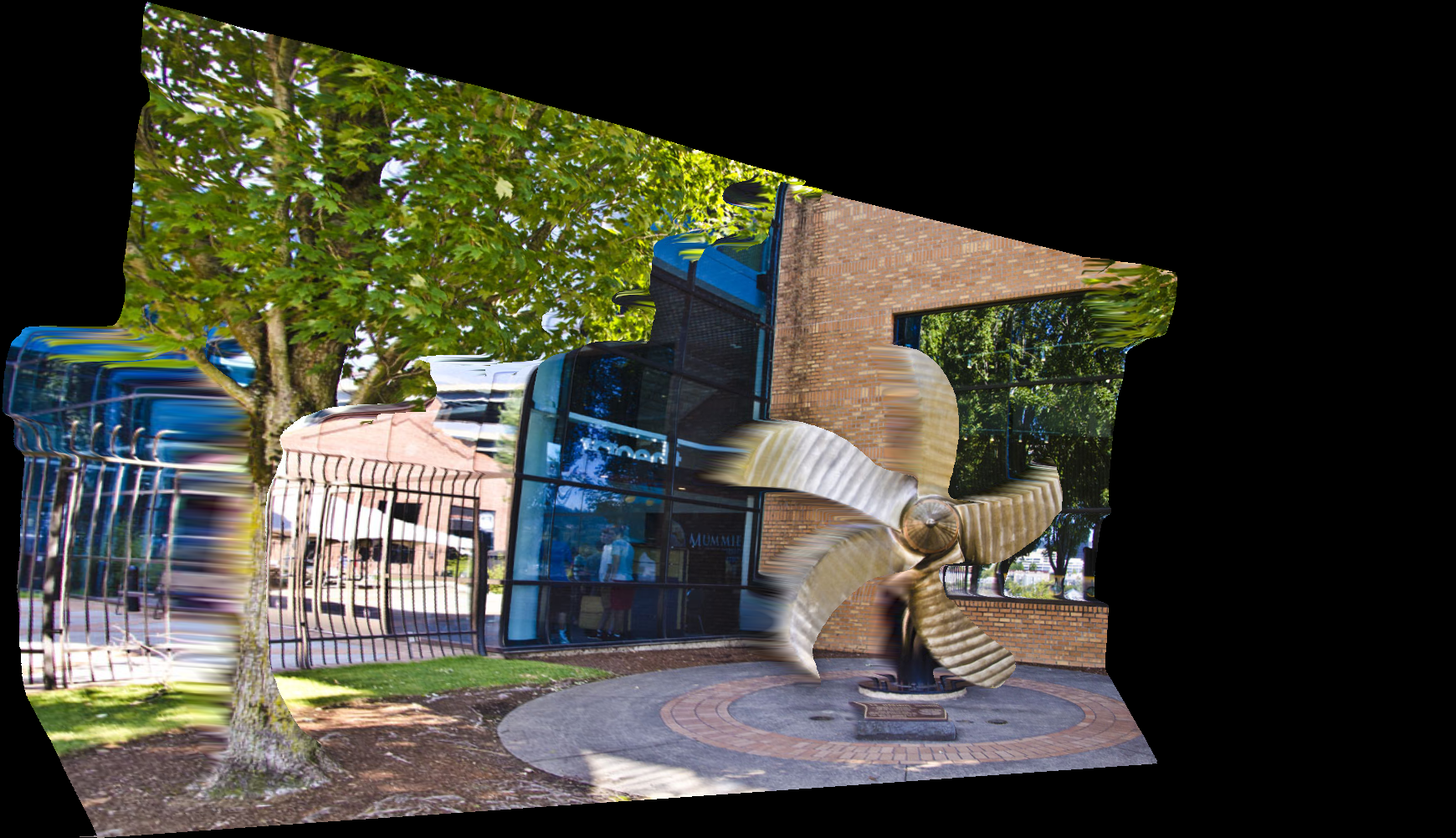}}
	\subfloat[Final warping result]{
		\includegraphics[width=0.32\textwidth]{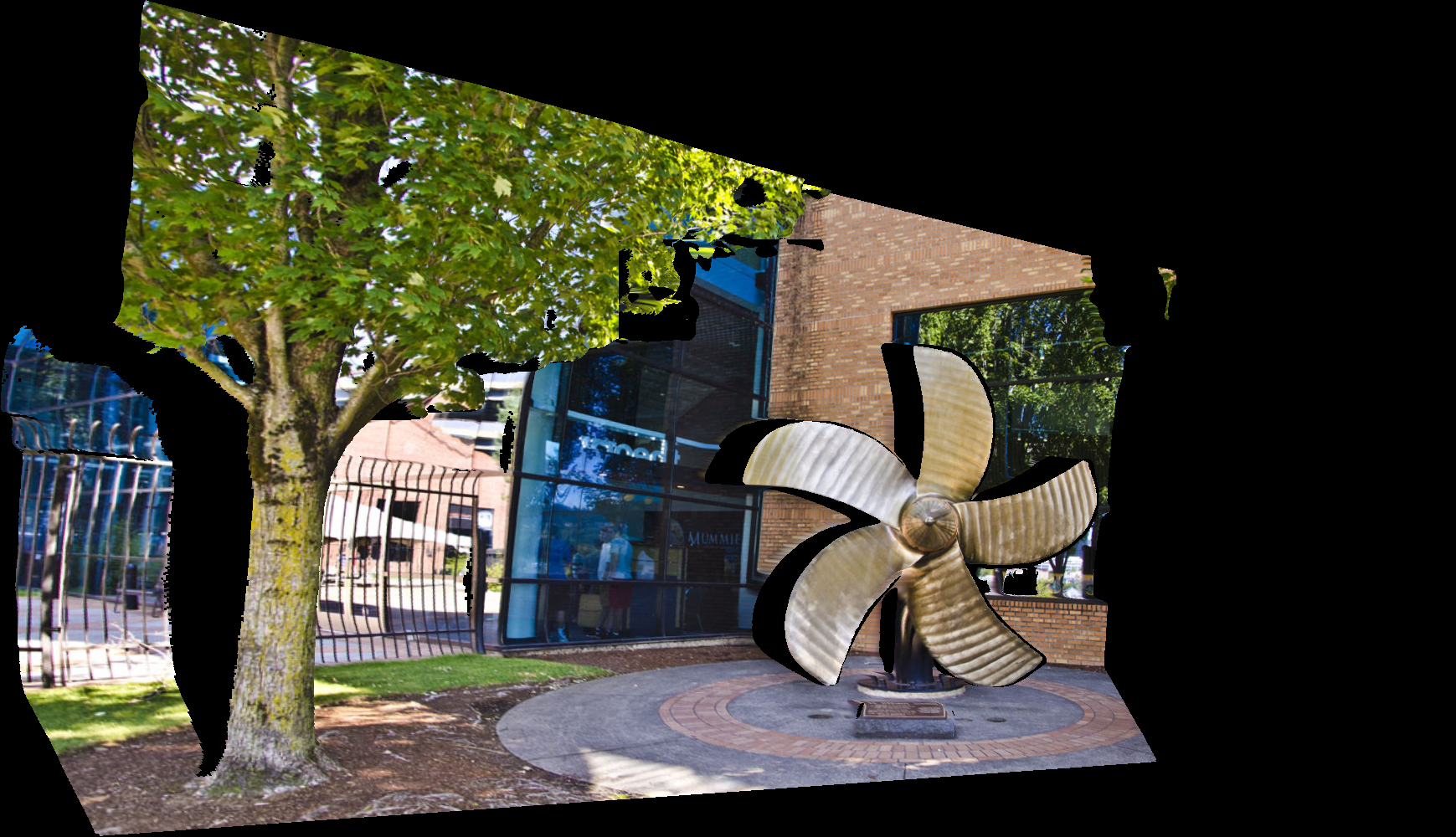}}\\
	\caption{Comparison of different image warping strategies on one test case. For clarity, we only draw 20$\times$30 mesh grids in (b) and (c).}
	\label{fig:warping}
\end{figure*}

\subsubsection{Backward mapping}
With the estimated $\mathrm{H}_{\infty}$, $\mathbf{e}'$, and the depth map of $I_t$, we can construct the pixel-to-pixel correspondences via Eq. (\ref{Hande}) and generate the warped target image using direct forward mapping. Fig. \ref{fig:warping}(a,b) shows the input images, and (d) shows the direct forward mapping result. Large voids represent that these regions are occluded in the target view and recur in the reference view. Forward mapping can generate content-consistent and view-consistent results, but it has floating coordinate issues, resulting in incomplete warped results. We propose to render the warped result via backward mapping. Specifically, we partition the target image into $C_1\times C_2$ regular grids and calculate the corresponding grid vertices via Eq. (\ref{Hande}), as shown in Fig. \ref{fig:warping}(b,c). Then we render each warped grid via backward bilinear interpolation and generate the complete warped images. Fig. \ref{fig:warping}(e) shows the warped result via backward mapping. Note that the backward mapping generates a total void-free result, but the occluded regions are wrongly inpainted by bilinear interpolation. 

\subsubsection{Void region recovery}
\label{subsubsec:occlusion}

To further improve the warped result, we introduce a void recovery algorithm to find wrongly inpainted regions generated by backward mapping and recover them to the void.

For each pixel $\mathbf{s}'$ in the overlapping region of the warped image domain, we map it back to the target image domain via estimating the infinity homography $\mathrm{H}^r_\infty$ and epipole $\mathbf{e'}^r$ from $I_r$ to $I_t$ using the feature matches $\{(\mathbf{q}_i, \mathbf{p}_i)\}_{i=1}^N$ and the depth map $z'=z'(\mathbf{x}')$ of $I_r$. The mapped pixel $\hat{\mathbf{s}}$ is computed as
\begin{equation}
	\hat{\mathbf{s}}=\pi\left(\mathrm{H}^r_\infty \tilde{\mathbf{s}}' + \mathbf{e'}^r/z'(\mathbf{s}')\right)
    \label{eq:He_2-1}
\end{equation}
$z'(\mathbf{s}')$ is the depth value of $\mathbf{s}'$ in the reference image. Then we calculate a bidirectional warping distance for $\mathbf{s}'$ as
\begin{equation}
	d(\mathbf{s}')=\|\mathbf{s}-\hat{\mathbf{s}}\|
\end{equation}
where $\mathbf{s}$ is the pixel coordinate in the target image domain computed by backward bilinear interpolation on the grid containing $\mathbf{s}'$.
The final warped target image $I_w$ is generated as
\begin{equation}
	I_w(\mathbf{s}')=\begin{cases}
		I_t(\mathbf{s}),&\mbox{if $d(\mathbf{s}')<\beta$}\\
		0,&\mbox{else}
	\end{cases}
\end{equation}
where $\beta$ is set as $1\%$ of the diagonal length of $I_t$. This means that if the bidirectional warping distance of a pixel is too large, it is most likely located in the void region, thus its intensity is set to 0.

For each warped grid $\mathbf{g}'_i$ in the non-overlapping region, we compute the horizontal and vertical length of $\mathbf{g}'_i$ and consider this grid is occluded in the target view and set $I_w(\mathbf{g}'_i)=0$ if its horizontal or vertical length is larger than two times of the average horizontal or vertical length of the whole warped grids.

\subsubsection{Overlapped pixels refining}
\label{subsubsec:overlap}

Besides, some other warped grids may overlap each other due to the estimation error of the epipolar geometry and depth map. We handle it as follows,
\begin{itemize}
	\item for two pixels $\mathbf{s}'_1, \mathbf{s}'_2$ in the overlapping region with the same coordinate $(x',y')$ in the warped image domain, we choose the one with the smaller photometric error, that is
	\begin{equation}
		I_w(x',y')=\begin{cases}
			I_t(\mathbf{s}_1),&\mbox{if $E(\mathbf{s}'_1)<E(\mathbf{s}'_2)$}\\
			I_t(\mathbf{s}_2),&\mbox{if $E(\mathbf{s}'_1)\geq E(\mathbf{s}'_2)$}
		\end{cases}
	\end{equation}
	where $\mathbf{s}_1, \mathbf{s}_2$ is the corresponding bilinear interpolated pixel coordinate of $\mathbf{s}'_1$ and $\mathbf{s}'_2$ in the target image domain, respectively. $E(\mathbf{s}'_1)$ (or $\mathbf{s}'_2$) is defined as
	\begin{equation}
		E(\mathbf{s}'_1)=\|I_r(\mathbf{s}'_1)-I_t(\mathbf{s}_1)\|
	\end{equation}
	
	\item for two pixels $\mathbf{s}'_1, \mathbf{s}'_2$ in the non-overlapping region with the same coordinate $(x',y')$ in the warped image domain, we choose the one with the smaller depth value, that is
	\begin{equation}
		I_w(x',y')=\begin{cases}
			I_t(\mathbf{s}_1),&\mbox{if $z(\mathbf{s}_1)<z(\mathbf{s}_2)$}\\
			I_t(\mathbf{s}_2),&\mbox{if $z(\mathbf{s}_1)\geq z(\mathbf{s}_2)$}
		\end{cases}
	\end{equation}
\end{itemize}
Fig. \ref{fig:warping}(f) shows our final warped result, using backward mapping, void recovery, and overlapped pixels refining can generate complete results with correct voids in both overlapping and non-overlapping regions. We further conduct a comprehensive experiment to validate the superiority of our image warping algorithm, which is demonstrated in Sec. \ref{subsec:ablation}. 

To clarify the optimal warping algorithm, we summarize the above three processes into Algorithm \ref{algor_2}, where the used symbols are listed in Table \ref{tab:notation}.

\begin{table}
    \centering
    \caption{Summary of the notation used in Algorithm \ref{algor_2}.}
    \label{tab:notation}
    \resizebox{\linewidth}{!}{
    \begin{tabular}{ll}
    \toprule
    Symbol & Description\\
    \midrule
        $O_w$ ($O_w^c$) & overlapping (non-overlapping) region in $I_w$\\
        $\mathrm{Bi}$ & bilinear interpolation map from $I_w$ to $I_t$\\
        $g$ & epipolar geometry map from $I_w$ to $I_t$\\
        $\mathrm{h}(\mathbf{x})$ ($\mathrm{v}(\mathbf{x})$) & horizontal (vertical) length of the mesh containing point $\mathbf{x}$\\
        $\ell_\mathrm{h}$ ($\ell_\mathrm{v}$) & average horizontal (vertical) length of all meshes in $I_w$\\
        \bottomrule
    \end{tabular}}
\end{table}

\begin{algorithm}
    \caption{optimal warping with void region recovery and overlapped pixel refinement}
    \label{algor_2}
        \KwIn{target image $I_t$, reference image $I_r$} 
        \KwOut{warped target image $I_w$}
        \tcp{void region recovery}
        \For{$\mathbf{s}'\in O_w (\mathbf{s}'\in O^c_w)$}{$d(\mathbf{s}')=\|\mathrm{Bi}(\mathbf{s}')-g(\mathbf{s}')\|$\\
        \eIf{$d(\mathbf{s}')<\beta$ $(\mathrm{h}(\mathbf{s}')\leq2\ell_{\mathrm{h}} \text{ or } \mathrm{v}(\mathbf{s}')\leq2\ell_{\mathrm{v}})$}
        {$I_w(\mathbf{s}')=I_t(\mathrm{Bi}(\mathbf{s}'))$}
        {$I_w(\mathbf{s}')=0$}
        } 
        \tcp{overlapped pixel refinement}
        \For{$\mathbf{s}'_1=\mathbf{s}'_2=(x',y')\in O_w$}
        {
        \eIf{$\|I_r(\mathbf{s}'_1)-I_t(\mathbf{s}'_1)\|<\|I_r(\mathbf{s}'_2)-I_t(\mathbf{s}'_2)\|$}
        {$I_w(x',y')=I_t(\mathrm{Bi}(\mathbf{s}'_1))$}
        {$I_w(x',y')=I_t(\mathrm{Bi}(\mathbf{s}'_2))$}
        }
        \For{$\mathbf{s}'_1=\mathbf{s}'_2=(x',y')\in O^c_w$}
        {
        \eIf{$z(\mathrm{Bi}(\mathbf{s}'_1))<z(\mathrm{Bi}(\mathbf{s}'_2))$}
        {$I_w(x',y')=I_t(\mathrm{Bi}(\mathbf{s}'_1))$}
        {$I_w(x',y')=I_t(\mathrm{Bi}(\mathbf{s}'_2))$}
        }
\end{algorithm}

\subsubsection{Void areas handling}
\label{subsubsec:void}


Noticing that there are voids in the final warped target image. It represents that the image contents are invisible in the target view and recur in the reference view. Usually, there are two kinds of voids, one in the overlapping region and the other in the non-overlapping region, as shown in Fig. \ref{fig:warping}(f). The voids in the overlapping region will be filled or overlapped by the reference image content after image blending. However, the voids in the non-overlapping region cannot be overlapped since there is no reference content here. Directly using bilinear interpolation to fill the large voids will result in severe artifacts, as shown in Fig. \ref{fig:warping}(e). Thus, after generating a panorama result via average blending, we apply the image inpainting method LaMa \cite{suvorov2022resolution} as a post-processing step to fill the voids in the panorama. Fig. \ref{fig:pipeline} and \ref{fig:comp} show several final inpainted stitching results. The voids can be plausibly inpainted.  

We summarize our image stitching method using depth maps in Algorithm \ref{algor_3}.

\begin{algorithm}
	\caption{Natural image stitching using depth maps}
	\label{algor_3} 
		\KwIn{$I_t$, $I_r$ and depth maps of the two images}
		\KwOut{final panorama result}
		estimating $\mathrm{H}_\infty$ and $\mathbf{e'}$ via Algorithm \ref{algor_1};\\
		constructing pixel-to-pixel correspondences via Eq. (\ref{Hande});\\
		generating warped target image via Algorithm \ref{algor_2};\\
		generating panorama result via simple average blending;\\
		inpainting void areas in the non-overlapping region to generate the final panorama result via the method LaMa \cite{suvorov2022resolution}.
\end{algorithm}

\section{Experiments}
\label{sec:exp}

A series of comparison experiments is conducted on three challenging datasets, Parallax \cite{zhang2014parallax}, SEAGULL \cite{lin2016seagull}, and MR \cite{herrmann2018robust}. The depth maps of images are estimated via the depth anything model \cite{yang2024depth}, $C_1\times C_2$ is set to the resolution of the input image to ensure pixel-level alignment. We compare our method with the state-of-the-art methods, including global homography (Homo), APAP \cite{zaragoza2014projective}, SPHP \cite{chang2014shape}, ANAP \cite{lin2015adaptive}, GSP \cite{chen2016natural}, REW \cite{li2018parallax}, SPW \cite{liao2020Single}, TFA \cite{li2019local}, LPC \cite{jia2021Leveraging}, UDIS++ \cite{nie2023parallax} and MHW \cite{liao2025parallax}. The parameters of existing methods are set as suggested by the original papers. To highlight the accuracy of image alignment, all stitching results are generated via simple average blending. 

\begin{table*}
\centering
		\caption{Quantitative comparisons between SOTA warping methods on different challenge datasets. DunHuangStitch resizes all the images to 512$\times$512 for stitching. The best is marked in \textbf{bold} and the second best is in \underline{underline}.}
        \resizebox{\linewidth}{!}{
		\begin{tabular}{l|ccc|ccc|ccc}
			\toprule
			& \multicolumn{3}{c|}{Parallax \cite{zhang2014parallax}} & \multicolumn{3}{c|}{SEAGULL \cite{lin2016seagull}} & \multicolumn{3}{c}{MR \cite{herrmann2018robust}}\\
			Method   & PSNR $\uparrow$ & SSIM $\uparrow$  & LPIPS $\downarrow$ & PSNR $\uparrow$  & SSIM $\uparrow$  & LPIPS $\downarrow$ & PSNR $\uparrow$  & SSIM $\uparrow$  & LPIPS $\downarrow$ \\
			\midrule
			Homo & 15.37 & 0.614 & 0.297 & 15.93 & 0.592 & 0.313 & 14.57 & 0.565 & 0.325 \\
			APAP \cite{zaragoza2014projective} & 16.84 & 0.659 & 0.237 & 16.63 & 0.613 & 0.282 & 15.89 & 0.608 & 0.282 \\
			SPHP \cite{chang2014shape} & 15.77 & 0.662 & 0.317 & 15.90 & 0.640 & 0.339 & 15.02 & 0.664 & 0.364 \\
			ANAP \cite{lin2015adaptive} & 16.87 & 0.666 & 0.227 & 17.00 & 0.640 & 0.247 & 16.11 & 0.619 & 0.269 \\
			GSP \cite{chen2016natural} & 17.30 & 0.698 & 0.215 & 17.34 & 0.676 & 0.228 & 15.28 & 0.621 & 0.274 \\
			REW \cite{li2018parallax} & 17.37 & 0.700 & 0.233 & 16.78 & 0.650 & 0.290 & 14.78 & 0.587 & 0.339\\
			SPW \cite{liao2020Single} & 16.33 & 0.642 & 0.250 & 16.49 & 0.602 & 0.285 & 15.49 & 0.574 & 0.291 \\
			TFA \cite{li2019local} & 16.00 & 0.637 & 0.319 & 16.63 & 0.650 & 0.293 & 14.45 & 0.567 & 0.366 \\
			LPC \cite{jia2021Leveraging} & 16.33 & 0.634 & 0.256 & 16.03 & 0.588 & 0.299 & 14.29 & 0.514 & 0.342 \\
			UDIS++ \cite{nie2023parallax} & 15.64 & 0.606 & 0.269 & 16.09 & 0.577 & 0.294 & 15.02 & 0.542 & 0.311 \\
            DunHuangStitch*~\cite{mei2024dunhuangstitch} & 17.32 & 0.652 & 0.215 & 17.65 & 0.642 & 0.235 & 16.38 & 0.618 & 0.259\\
			MHW \cite{liao2025parallax} & \underline{19.19} & \underline{0.752} & \underline{0.192} & \underline{18.69} & \underline{0.713} & \underline{0.226} & \underline{18.55} & \underline{0.736} & \underline{0.197} \\
			\rowcolor{gray!30}Ours  & \textbf{20.34} & \textbf{0.780} & \textbf{0.146} & \textbf{21.19} & \textbf{0.780} & \textbf{0.141} & \textbf{21.55} & \textbf{0.822} & \textbf{0.099} \\
			\bottomrule
		\end{tabular}
        }
        \label{tab:quality}%
\end{table*}

\subsection{Quantitative comparison}

To accurately evaluate the performance of our method, we introduce three metrics, PSNR, SSIM \cite{wang2004image}, and LPIPS \cite{zhang2018unreasonable} to evaluate the alignment quality and compare with other methods. The three metrics are calculated based on the overlapping regions of warped images. 

We evaluate the whole results based on the metrics and calculate the average PSNR, SSIM, and LPIPS scores, as shown in Table \ref{tab:quality}. In rare test cases, Homo or ANAP \cite{lin2015adaptive} fail to stitch the images reasonably, we omit such cases in their average metrics calculation. The global homography (Homo) and warping models aiming to alleviate the distortion in the non-overlapping region (SPHP, SPW, LPC) are not able to handle the large parallax and eliminate local structure misalignments, such that receives worse scores. GSP and REW could achieve better alignment quality and hence get better scores. By segmenting images into contents and aligning them separately, MHW achieves the second-best score. Among all the tested methods, our proposed method achieves the best scores by a large margin, improving by up to 16.17\% in PSNR, 11.68\% in SSIM, and 49.75\% in LPIPS.

\subsection{Qualitative comparison}

\begin{figure*}
	\centering
    \includegraphics[width=\textwidth]{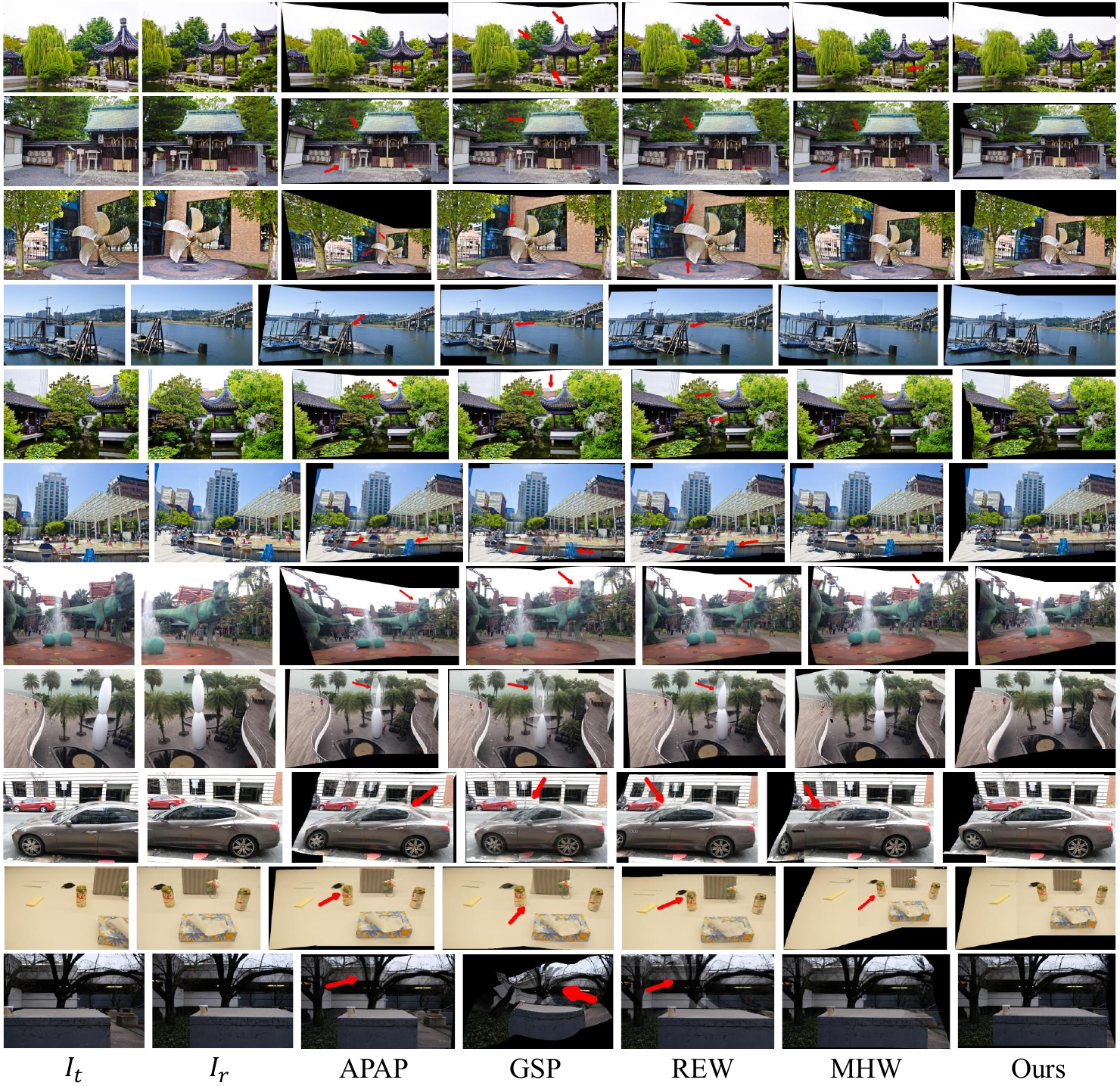}
	\caption{Comparison of the image stitching results obtained by our method with that of the four state-of-the-art existing methods: APAP \cite{zaragoza2014projective}, GSP \cite{chen2016natural}, REW \cite{li2018parallax}, and MHW \cite{liao2025parallax} (best view in color and zoom in).}
	\label{fig:comp}
\end{figure*}

We also compare the visual results qualitatively. Fig.~\ref{fig:comp} presents several comparison results of our method with other state-of-the-art methods on the three datasets. Each case contains large parallax and drastically varying depths. The existing methods suffer from ghosting effects in various parts, which are indicated by red arrows. With the help of the depth map, our warping model can accurately align the images, hence outperforming all the other methods visually.

\subsection{Ablation study}\label{subsec:ablation}

We validate the effectiveness of every module in our method by evaluating the average metrics on the three datasets, as shown in Table \ref{tab:ablation}.

\begin{table*}
	\centering
		\caption{Ablation studies on different challenge datasets. The best is marked in \textbf{bold} and the second best is in \underline{underline}.}
        \resizebox{\linewidth}{!}{
		\begin{tabular}{l|l|ccc|ccc|ccc}
			\toprule
			& & \multicolumn{3}{c|}{Parallax \cite{zhang2014parallax}} & \multicolumn{3}{c|}{SEAGULL \cite{lin2016seagull}} & \multicolumn{3}{c}{MR \cite{herrmann2018robust}}\\
			& Method   & PSNR $\uparrow$ & SSIM $\uparrow$  & LPIPS $\downarrow$ & PSNR $\uparrow$  & SSIM $\uparrow$  & LPIPS $\downarrow$ & PSNR $\uparrow$  & SSIM $\uparrow$  & LPIPS $\downarrow$ \\
			\midrule
			1 & homo-RANSAC & 17.71 & 0.704 & 0.199 & 18.61 & 0.716  & 0.196  & 20.28 & 0.770 & 0.160\\
			2 & multi-RANSAC & 18.75 & 0.730  & 0.175 & 18.64 & 0.708 &  0.184 & 19.18 & 0.730 & 0.161 \\
			\midrule
			3 & direct forward & 18.37 & \textbf{0.783} & 0.155 & 18.61 & \textbf{0.809} & \underline{0.146} & 18.78 & 0.778 & 0.152 \\
			4 & direct backward & 16.93 & 0.663 & 0.254 & 16.82 & 0.636 & 0.273 & 16.84 & 0.615  &  0.229 \\
			5 & backward + void & 19.82 & 0.767 & 0.154 & \underline{20.81} & 0.766  & 0.147  & \underline{21.24} & \underline{0.813} &  \underline{0.104} \\
			\midrule
			6 & Ours (\textbf{ViT-S}) & 19.77 & 0.763 & 0.154 & 19.96 & 0.761 & 0.155 & 21.00 & 0.803 & 0.109\\
			7 & Ours (\textbf{ViT-B}) & \underline{20.06} & 0.767 & \underline{0.149} & 20.31 & 0.762 & 0.150 & 21.17 & 0.801 & 0.108\\
			\rowcolor{gray!30}8 & Ours (\textbf{ViT-L}) & \textbf{20.34} & \underline{0.780} & \textbf{0.146} & \textbf{21.19} & \underline{0.780} & \textbf{0.141} & \textbf{21.55} & \textbf{0.822} & \textbf{0.099} \\
			\bottomrule
		\end{tabular}
        }
        \label{tab:ablation}
\end{table*}

\begin{table*}
	\centering
		\caption{Comparison of elapsed time (seconds) for different datasets. UDIS++ is tested with GPU acceleration, DunHuangStitch resizes all the images to 512$\times$512 for stitching.}
		\label{tab:times}%
		\begin{tabular}{lccc}
			\toprule
			   & Parallax \cite{zhang2014parallax} & SEAGULL \cite{lin2016seagull} & MR \cite{herrmann2018robust} \\
			\midrule
			Resolution & 696$\times$1028 & 707$\times$960 & 408$\times$569 \\
            \midrule
			APAP~\cite{zaragoza2014projective} & 5.444 & 4.522 & 2.275 \\
			SPHP~\cite{chang2014shape} & 9.977 & 4.551 & 2.076\\
			ANAP~\cite{lin2015adaptive} & 18.265 & 17.724 & 8.010\\
                REW~\cite{li2018parallax} & 3.501 & 2.935 & 1.314\\
                SPW~\cite{liao2020Single} & 40.753 & 41.473 & 3.545 \\
                TFA~\cite{li2019local} & 7.426  & 5.755 & 2.672 \\
                LPC~\cite{jia2021Leveraging} & 26.369 & 29.885 & 3.112\\
                MHW~\cite{liao2025parallax} & 7.903 & 6.435 & 2.408\\
                UDIS++~\cite{nie2023parallax} & 0.514  &  0.304 & 0.278\\
             DunHuangStitch*~\cite{mei2024dunhuangstitch} & 0.100  &  0.111 & 0.129\\
                \midrule
               \rowcolor{gray!30} Ours & 4.151 &  3.627 & 1.040\\
                ~~1. SIFT~\cite{lowe2004distinctive} & 2.278 & 1.581 & 0.427\\
                ~~2. depth-RANSAC & 0.029 & 0.051 & 0.030\\
                ~~3. epipolar estimation & 0.244 & 0.223 & 0.090\\
                ~~4. optimal warping & 1.927 & 1.539 & 0.522\\
			\bottomrule
		\end{tabular}%
\end{table*}

\subsubsection{Robust fitting}

We integrate different robust fitting methods, including homography-based RANSAC (homo-RANSAC), multiple sampling RANSAC (multi-RA-NSAC), and our depth-based RANSAC, into the epipolar geometry estimation and optimal image warping to generate aligned results and evaluate the three metrics, as shown in experiments 1,2,8 of Table \ref{tab:ablation}. The homo-RANSAC cannot identify sufficient matched features for large parallax cases, thus providing the lowest alignment accuracy. Although the multi-RANSAC identified sufficient features as our depth-RANSAC, it has a lower accuracy than ours. We believe the reason is that the multi-RANSAC has the worst mapping error (see Fig. \ref{fig:feature}) such that the subsequent optimal image warping cannot alleviate it.

\subsubsection{Optimal image warping}

We quantitatively evaluate the alignment quality of different warping strategies, including direct forward mapping, direct backward mapping, backward mapping with void recovery but without overlapped pixels refining (``backward + void''), and our final warping, as shown in experiments 3-5,8 of Table \ref{tab:ablation}. With all the warping modules included, our final warping produces the warped result with much better alignment quality than the other three strategies.

\subsubsection{Depth estimation}

Since the depth anything model \cite{yang2024depth} provides three pre-trained models of varying scales for robust relative depth estimation: small (\textbf{ViT-S}), base (\textbf{ViT-B}), and large (\textbf{ViT-L}). We compare the alignment accuracy of our method when using different pre-trained models to estimate the depth maps, as shown in experiments 6-8 of Table \ref{tab:ablation}. The larger scale of the pre-trained model leads to more accurate estimated depth maps, and thus better alignment accuracy. Moreover, the alignment accuracy does not drop much from ViT-L to ViT-S, which highlights the robustness of our approach, indicating that a pixel-wise precise depth map is not a strict requirement for its successful application.
Note that experiments 1-5 adopt the large scale of the pre-trained model (ViT-L) for a fair comparison.

\subsection{Evaluation on computational efficiency}

We also compare the computational efficiency of our method with other methods on the three datasets. All the experiments are run under the same hardware configurations, except that UDIS++ is tested with NVIDIA RTX 3090. Table \ref{tab:times} shows the average time of each method. 
Generally, the time cost increases as the resolution increases for most methods.
SPW and LPC take too much time as they involve line detection and matching. With GPU acceleration, the learning-based method UDIS++ takes the least time. Among all traditional methods, our method takes the second least time in the Parallax and SEAGULL datasets, and the least time in the MR dataset. 
The runtime performance of each module in the proposed framework is further evaluated, including SIFT detection and matching~\cite{lowe2004distinctive}, depth-RANSAC, epipolar geometry estimation ($\mathrm{H}_\infty$, $\mathbf{e}'$), and optimal warping. Experimental results show that feature detection and image warping dominate the overall runtime, while the processing times for depth-RANSAC and epipolar estimation remain negligible.
Overall, our method incurs a very small computational cost when aligning two images with large parallax.

\begin{figure*}[t]
	\centering
	\subfloat[Input images and depth maps]{
		\includegraphics[width=0.24\textwidth]{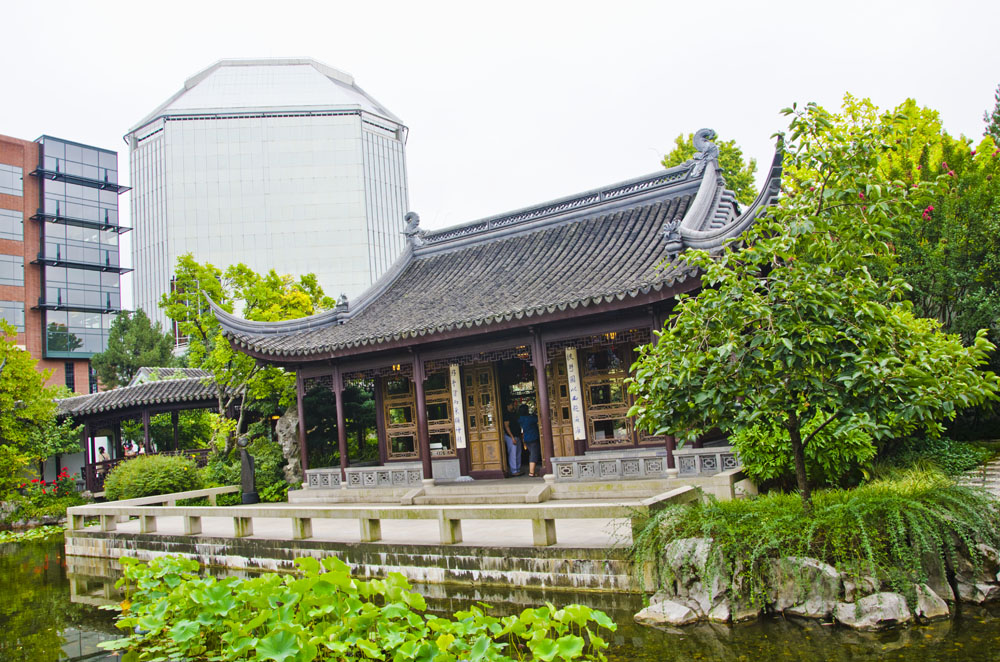}
		\includegraphics[width=0.24\textwidth]{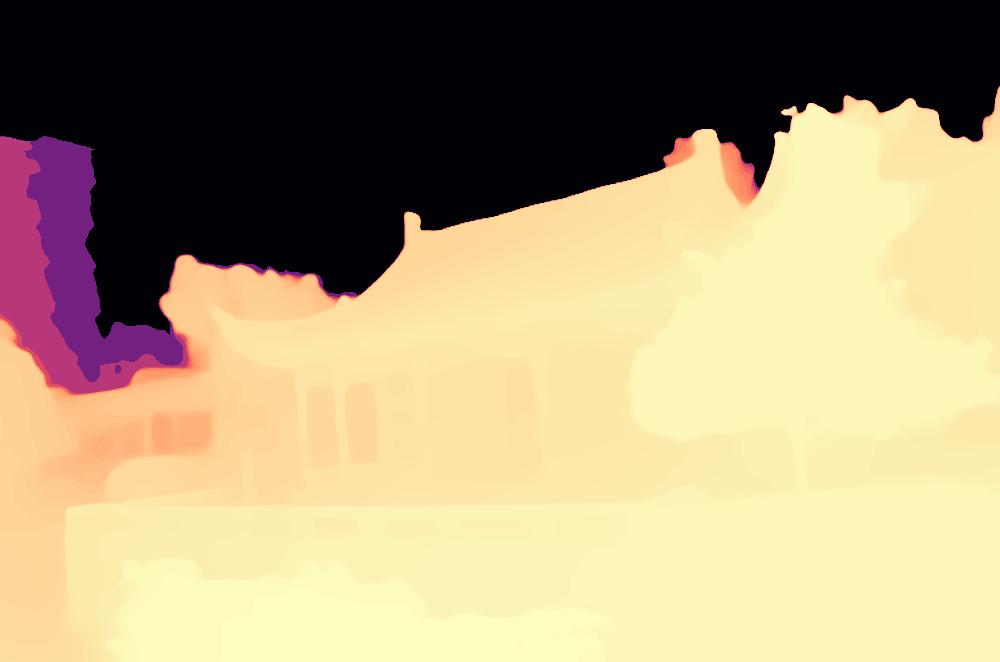}
		\includegraphics[width=0.24\textwidth]{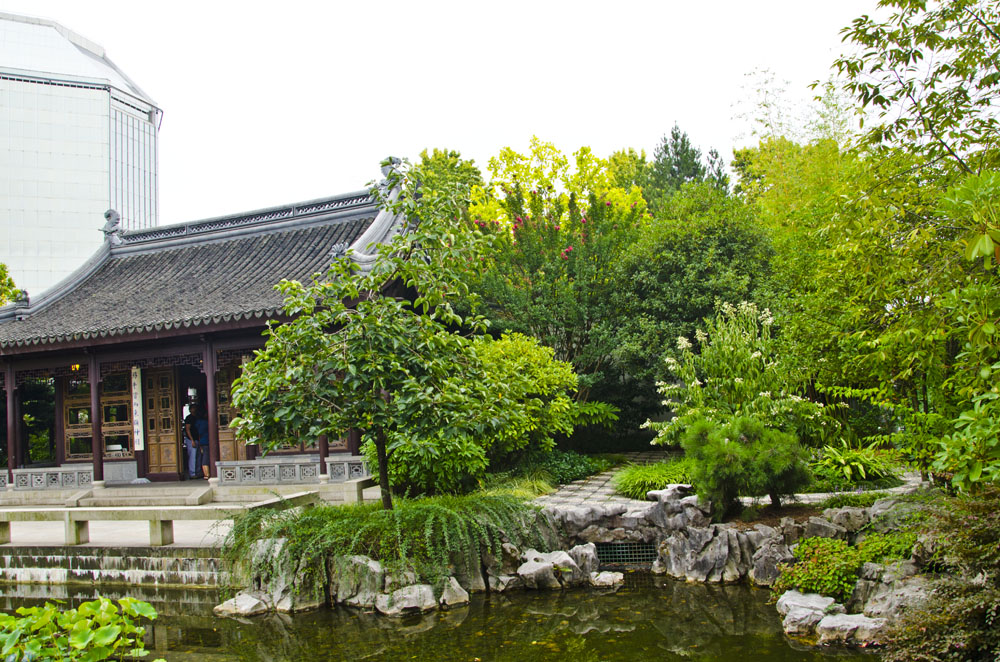}
		\includegraphics[width=0.24\textwidth]{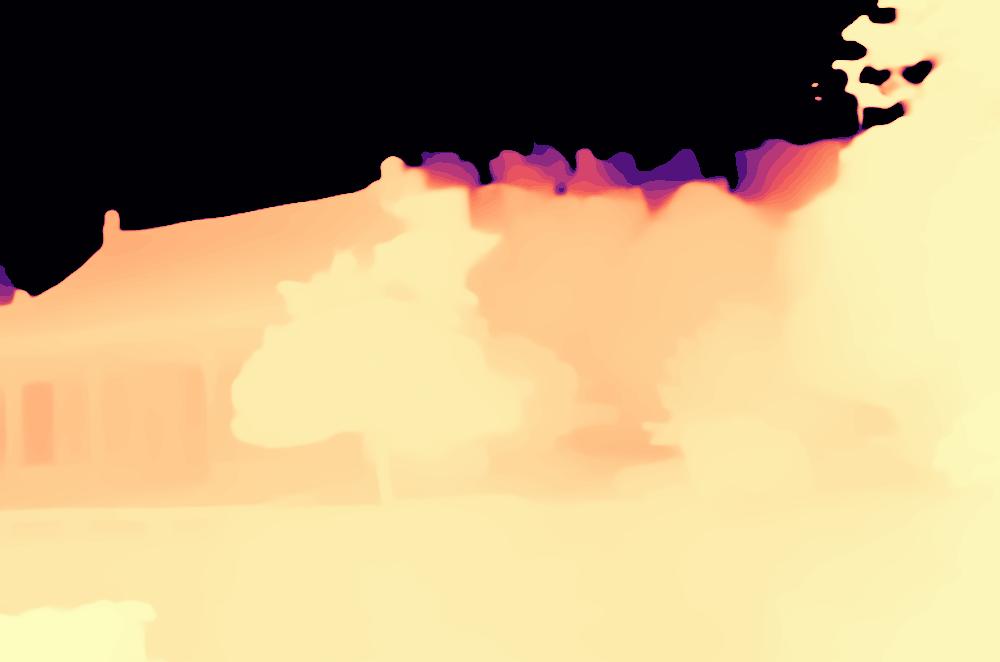}}\\
	\subfloat[Warped target image]{
		\includegraphics[width=0.487\textwidth]{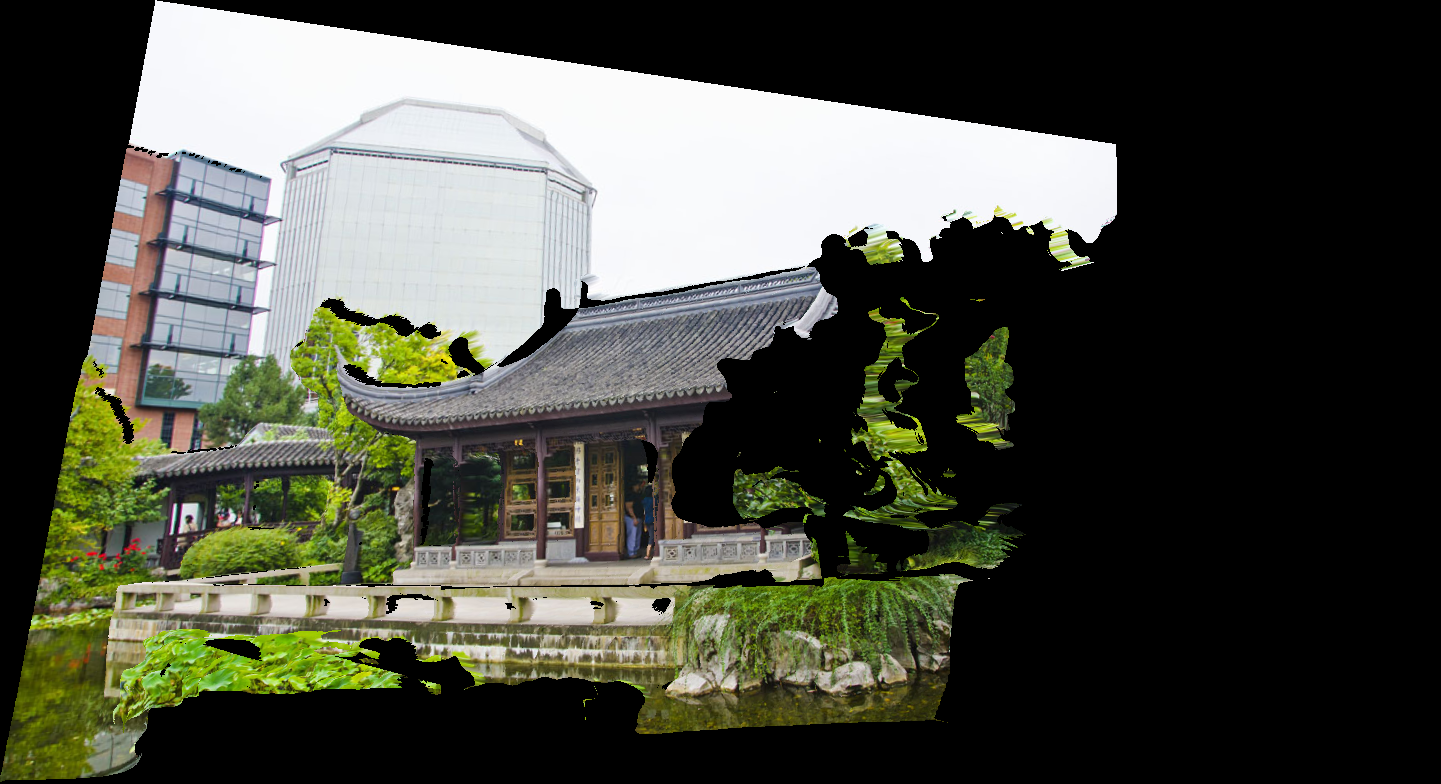}}
	\subfloat[Final panorama]{
		\includegraphics[width=0.487\textwidth]{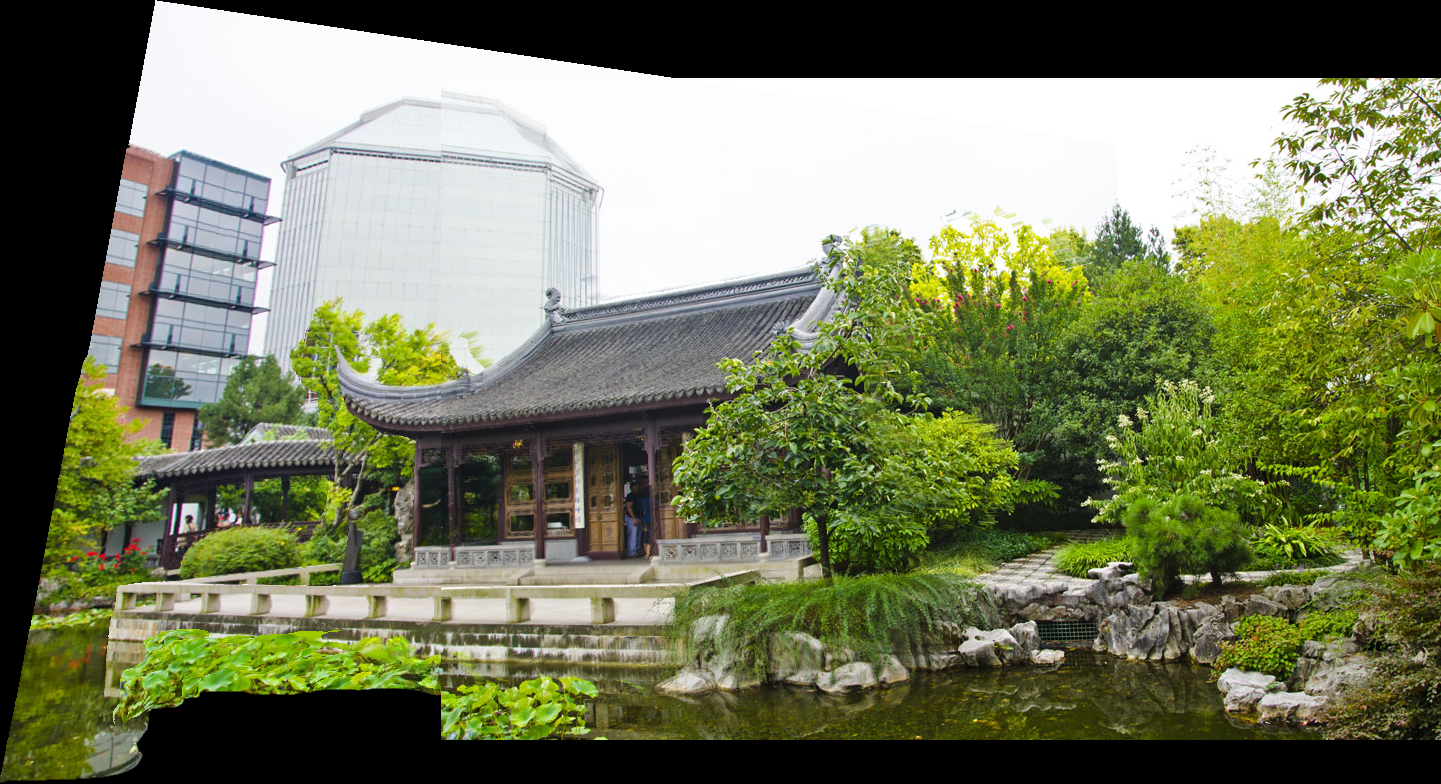}}\\
    \subfloat[Input images and depth maps]{
		\includegraphics[width=0.24\textwidth]{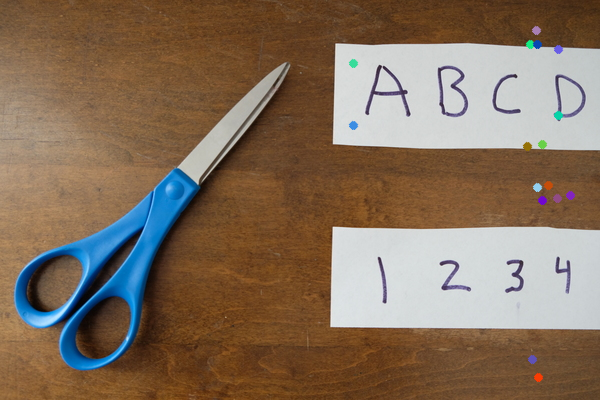}
		\includegraphics[width=0.24\textwidth]{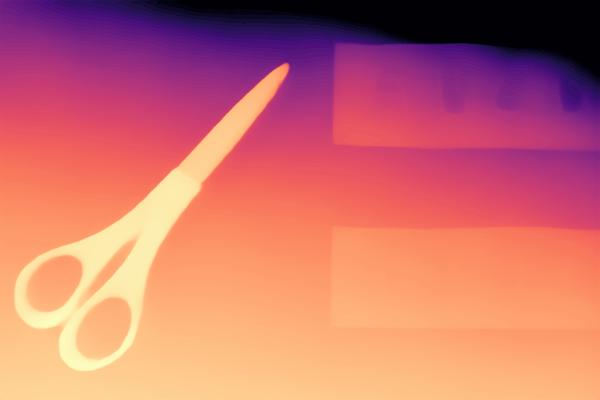}
		\includegraphics[width=0.24\textwidth]{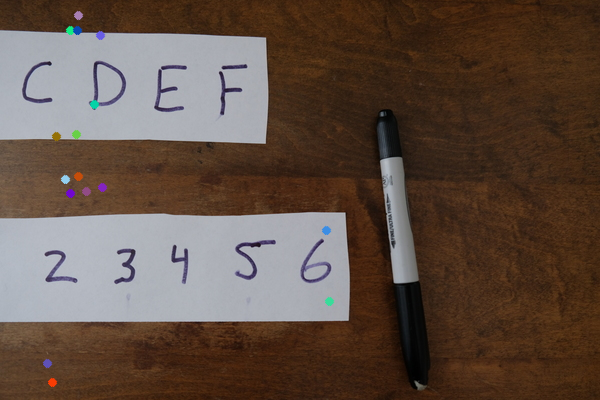}
		\includegraphics[width=0.24\textwidth]{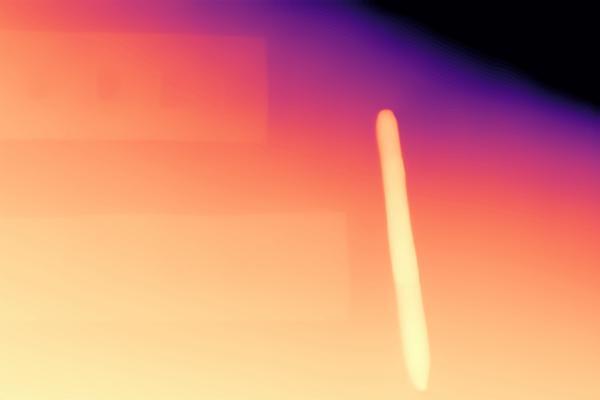}}\\
	\subfloat[Warped target image]{
		\includegraphics[width=0.487\textwidth]{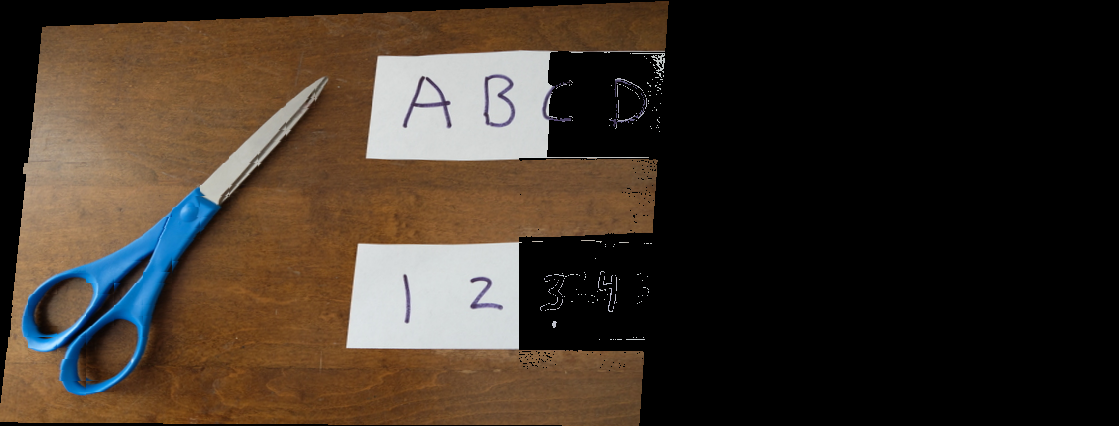}}
	\subfloat[Final panorama]{
		\includegraphics[width=0.487\textwidth]{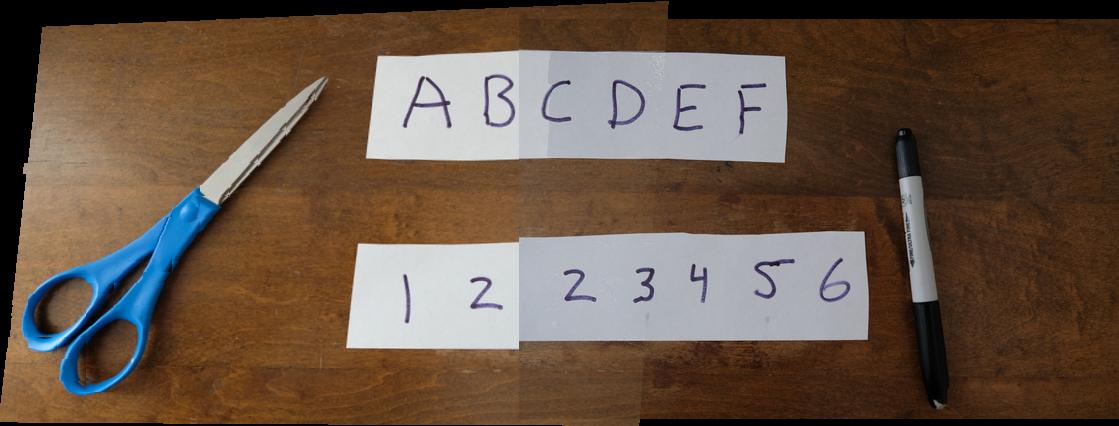}}\\    
	\caption{Failure examples of the proposed method.}
	\label{fig:failure}
\end{figure*}

\subsection{Limitation and Discussion}

In this paper, we validate that depth maps can help align images with large parallax. The more accurate the depth maps are, the better our method can perform. For images with low-quality estimated depth maps, our method may fail to calculate the correct infinite homography $\mathrm{H}_\infty$ and epipole $\mathbf{e}'$, thus cannot provide accurate alignment in the overlapping region and view-consistent result in the non-overlapping region. 
Fig.~\ref{fig:failure} shows one kind of failure example of our proposed method. 
When the estimated depth maps are unreliable, particularly in the overlapping region (the light cyan building), the proposed method may produce local misalignments, as shown in Fig. \ref{fig:failure}(c). This is because inaccurate depth causes incorrect epipolar geometry estimation between views. A potential solution is to introduce a residual warping strategy to refine alignment in these regions. Such residual corrections, learned or optimized after initial stitching, can compensate for small geometric inconsistencies caused by imperfect depth.

In scenes with limited structural or textural information, the scarcity of distinctive features can significantly degrade epipolar geometry estimation, a limitation also observed in traditional feature-based stitching methods. As illustrated in Fig. \ref{fig:failure}(d-f), such a failure case occurs when only a small number of reliable feature correspondences (marked in the input images) remain after the depth-RANSAC process. Although the estimated depth map appears reasonable, the resulting epipolar geometry is insufficient to correctly align low-texture regions. To enhance robustness in these scenarios, an adaptive geometry estimation module could be integrated, for example by leveraging learning-based epipolar geometry estimation to better handle texture-sparse scenes.

The current work focuses on pairwise image stitching, which is consistent with most existing stitching frameworks. However, the proposed method can be naturally extended to multiple image scenarios. By leveraging the pairwise alignment module within a global epipolar geometry estimation framework, it is possible to achieve consistent alignment across several overlapped views. Future work will explore this extension to enhance the applicability of the proposed method.

\section{Conclusions}
\label{sec:conclusion}

This paper proposes an image stitching method using depth maps. Our main contribution is to provide a method that leverages depth maps to address the challenge of parallax. Experimental results show that the proposed method provides the best \textit{content-consistent} alignment in the overlapping region and \textit{view-consistent} result in the non-overlapping region. Besides, it takes a very small computational cost. Future research includes reducing the dependence on the depth maps for the whole input image.
\appendix

\section*{Declaration of Competing Interest}
The authors declare that they have no known competing financial interests or personal relationships that could have appeared to influence the work reported in this paper.

\section*{Acknowledgements}

This work is partially supported by the Natural Science Foundation of Henan Province under Grant 222300420140, the National Natural Science Foundation of
China under Grant 12171324, the Guangdong Provincial Pearl River Talents Program under Grant 2021QN02X310, and the Guangdong Basic and Applied Basic Research Foundation under Grant 2024A1515010506.




  \bibliographystyle{elsarticle-num} 
  \bibliography{refs}



%
%
%
\end{document}